\PassOptionsToPackage{numbers, compress}{natbib}

\documentclass{article}
\usepackage[OT1]{fontenc}
\usepackage[table]{xcolor}
\usepackage{booktabs}
\usepackage{microtype}
\usepackage{graphicx}
\usepackage{subcaption}
\usepackage{booktabs} 
\usepackage{multirow}
\usepackage{wrapfig}
\usepackage{placeins}
\usepackage{titletoc}
\usepackage{hyperref}

\definecolor{darkgreen}{rgb}{0.0, 0.6, 0.0}
\definecolor{darkred}{rgb}{0.6, 0.0, 0.0}
\definecolor{lightyellow}{RGB}{255, 252, 230}

\usepackage[most]{tcolorbox}
\tcbset{
  colback=gray!6,    
  colframe=gray!100,  
  boxrule=0.8pt,
}

\usepackage[main, final]{neurips_2026}

\usepackage{amsmath}
\usepackage{amssymb}
\usepackage{mathtools}
\usepackage{amsthm}
\usepackage{enumitem}

\usepackage[capitalize,noabbrev]{cleveref}

\theoremstyle{plain}

\theoremstyle{definition}

\theoremstyle{remark}

\usepackage[textsize=tiny]{todonotes}

\title{ViT-AdaLA: Adapting Vision Transformers with Linear Attention}

\author{%
  Yifan Li$^1$\thanks{\textit{This work was done during an internship at Adobe Research.}} \quad
  Seunghyun Yoon$^2$ \quad
  Viet Dac Lai$^2$ \quad
  Franck Dernoncourt$^2$ \quad
  \quad \\ \bfseries
  Jason Kuen$^2$ \quad
  Yu Kong$^1$ \quad
  Trung Bui$^2$ \\
  $^1$Michigan State University, \texttt{\{liyifa11, yukong\}@msu.edu}\\
  $^2$Adobe Research, \texttt{\{bui, dernoco, kuen, daclai\}@adobe.com}\\
  \texttt{} \\
}

\begin{document}

\maketitle
\begin{abstract}
  Vision Transformers (ViTs) based vision foundation models (VFMs) have achieved remarkable performance across diverse vision tasks, but suffer from quadratic complexity that limits scalability to long sequences. Existing linear attention approaches for ViTs are typically trained from scratch, requiring substantial computational resources, while linearization-based methods developed for large language model decoders do not transfer well to ViTs. To address these challenges, we propose \textbf{ViT-AdaLA}, a novel framework for effectively adapting and transferring prior knowledge from VFMs to linear attention ViTs. ViT-AdaLA consists of three stages: attention alignment, feature alignment, and supervised fine-tuning. In the attention alignment stage, we align vanilla linear attention with the original softmax-based attention in each block to approximate the behavior of softmax attention. However, residual approximation errors inevitably accumulate across layers. We mitigate this by fine-tuning the linearized ViT to align its final-layer features with a frozen softmax VFM teacher. Finally, the adapted prior knowledge is transferred to downstream tasks through supervised fine-tuning. Extensive experiments on classification and segmentation tasks demonstrate the effectiveness and generality of ViT-AdaLA over various state-of-the-art linear attention counterpart. Project page is available: \url{https://jackyfl.github.io/vitadala-project-page}.
\end{abstract}
\vspace{-15pt}
\section{Introduction}

\begin{wrapfigure}{r}{0.6\textwidth}
    \centering
    \vspace{-10pt}
    \includegraphics[width=0.6\textwidth]{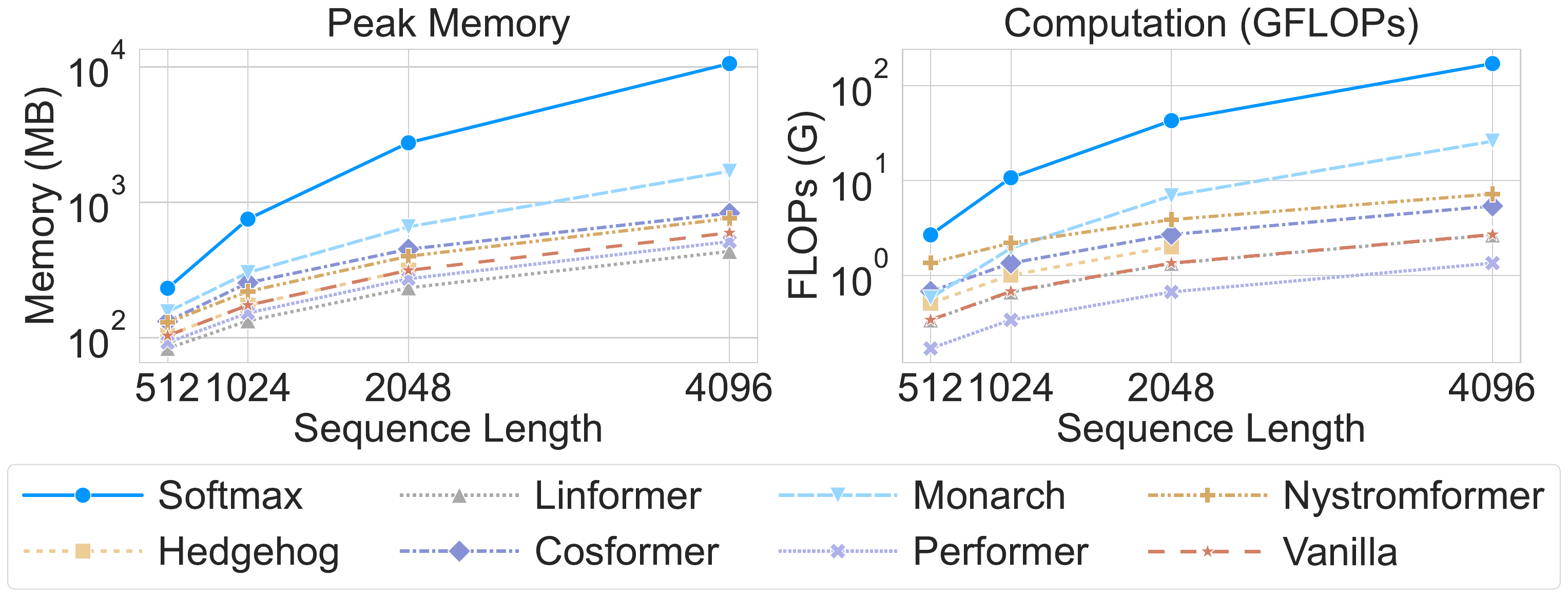}
    \caption{Efficiency comparison of different attentions, including peak memory and GFLOPS varying with sequence length. Only attention module is benchmarked in these experiments. ``Vanilla'' indicates the vanilla linear attention \cite{katharopoulos2020transformers}.}
    \label{fig:attention_overhead}
    \vspace{-10pt}
\end{wrapfigure}
Vision Transformers (ViTs) \cite{dosovitskiyimage} based vision foundation models (VFMs) such as DINOv2~\cite{oquab2024dinov2} and CLIP \cite{radford2021learning} have been widely adopted across a broad range of computer vision tasks \cite{Li_2025_ICCV}, including segmentation, detection, visual question answering (VQA), depth estimation, and image, video, and 3D point-cloud generation. However, the standard softmax-based self-attention in ViTs scales quadratically with the number of visual tokens, leading to substantial computational and memory overhead as the sequence length increases as shown in Fig.~\ref{fig:attention_overhead}. This limitation becomes increasingly acute as modern vision applications demand processing long-sequence visual tokens.

\begin{figure}[t]
    \centering

    \begin{subfigure}{0.34\textwidth}
        \centering
        \includegraphics[width=\linewidth]{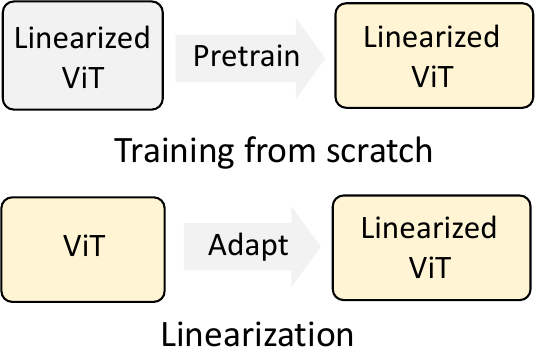}
        \caption{Linearization paradigm.}\label{fig:linear_attention_training_paradigm}
    \end{subfigure}
    \hfill
    \begin{subfigure}{0.62\textwidth}
        \centering
        \includegraphics[width=\linewidth]{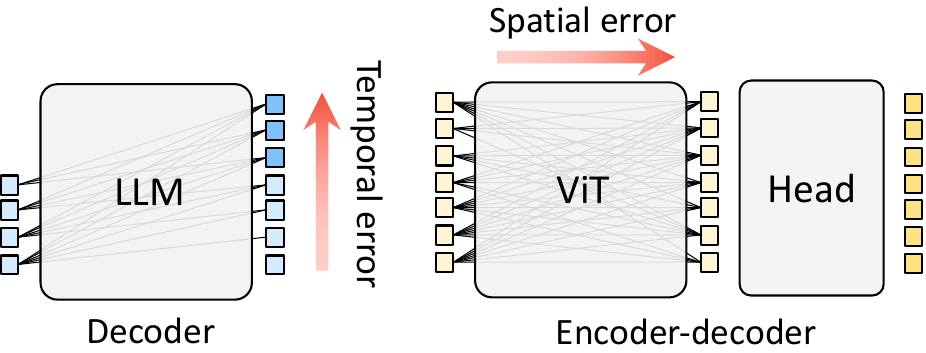}
        \caption{Comparison of decoder and encoder-decoder architecture.}
        \label{fig:enc-dec-comparison}
    \end{subfigure}

    \caption{(a) Comparison between training-from-scratch and linearization paradigms for ViTs with linear attention. Training-from-scratch linear attention paradigms focus on designing accurate attention approximation methods, which typically require large-scale pretraining to acquire strong prior knowledge. In contrast, ViT linearization leverages an off-the-shelf pretrained ViT, substantially reducing the need for extensive pretraining. (b) Comparison of decoder and encoder–decoder architectures. In decoder-based LLMs, the LLM serves as both the feature extractor and the target generator, where the error propagates temporally. In contrast, in vision models, ViTs function solely as feature extractors, while a separate task-specific head is responsible for target generation. The ViT error propagates spatially and hierarchically.}
    \label{fig:two_figures}
\end{figure}


To address the computational and memory bottlenecks of ViTs, extensive efforts have been devoted to improving the efficiency of softmax-based self-attention, including attention matrix optimization \cite{dao2022flashattention}, token reduction methods \cite{rao2021dynamicvit}, distillation~\cite{touvron2021training}, sliding-window mechanisms \cite{liu2021swin}, sequence modeling approaches \cite{gu2023mamba}, and linear attention variants~ \cite{katharopoulos2020transformers}. Among these, linear attention methods are particularly attractive, as they reduce the quadratic complexity ($\mathcal{O}(N^2 D)$), to linear complexity ($\mathcal{O}(ND^2)$), where $N$ and $D$ indicate the sequence length and the feature dimension, respectively. Existing linear attention approaches can be categorized into two types (Fig. \ref{fig:linear_attention_training_paradigm}): training from scratch~\cite{yaras2025monarchattention, xiong2021nystromformer} and linearization~\cite{zhanghedgehog, zhanglolcats, liu2025lawcat, lanliger, goldstein2025radlads}.
The former focuses on designing accurate softmax-approximation methods, and trains a linearized ViT entirely from scratch, typically requiring large-scale pretraining before fine-tuning on downstream tasks, especially for the VFMs designed as general-purpose feature extractors. Without such extensive pretraining, these approaches often suffer from severe performance degradation when directly adapted to downstream scenarios (see Tab. \ref{tab:imagenet_benchmarks}, \ref{tab:ade20k_bench}, \ref{tab:cityscapes_bench}), limiting their practicality under realistic data and compute constraints. The latter, in contrast, inherits prior knowledge from the softmax-based VFMs and therefore requires substantially fewer pretraining steps than training-from-scratch methods. 

However, existing work on linearization stream such as LoLCATS \cite{zhanglolcats} has primarily focused on large language models (LLMs), which are decoder-based transformers and differ fundamentally from encoder–decoder-based vision models (see Fig. \ref{fig:enc-dec-comparison}). In decoder-only LLMs, the model acts as both a feature extractor and a target generator, whereas in vision models, the ViT primarily serves as a feature extractor and a separate prediction head functions as the generator. Consequently, directly transferring linear attention adaptation paradigms from LLMs to ViTs leads to a substantial performance drop. We attribute this to divergent error propagation: while LLM errors accumulate temporally, ViT errors accumulate spatially and hierarchically. This distorts the global semantic manifold essential for dense prediction, making feature alignment non-negotiable to preserve the spatial consistency vision tasks require. More analyses are provided in Appendix \ref{sec:layer_sensitivity}, \ref{sec:spatia_error}.



\begin{figure*}[!t]
    \centering
    \includegraphics[width=0.98\linewidth]{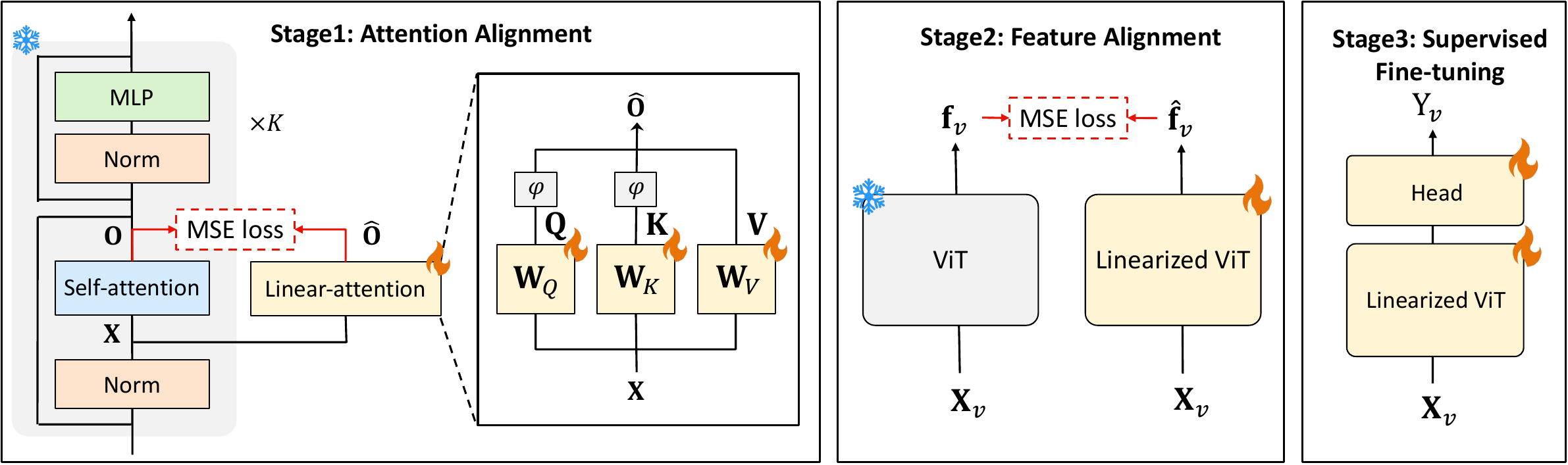}
    \caption{ViT-AdaLA consists of three stages: attention alignment, feature alignment, and supervised fine-tuning. First, softmax attention is approximated by tuning only the linear attention modules.  Second, to mitigate residual approximation errors that accumulate across layers, the feature alignment stage finetunes the entire linearized model by aligning its final-layer representations with those of the original softmax-based teacher. Finally, supervised fine-tuning is performed to transfer the adapted prior knowledge to downstream tasks.}
    \label{fig:ViT-AdaLA}
\end{figure*}

To address these challenges, we introduct \textbf{ViT-AdaLA}~(\textit{Adapting Vision Transformers with Linear Attention}). 
ViT-AdaLA consists of three stages designed to inherit knowledge from a pretrained softmax-based ViT and transfer it to downstream tasks: 
{attention alignment}, {feature alignment}, and {supervised fine-tuning}. To effectively adapt the prior knowledge from the VFMs, we first align the linear attention module with the original softmax attention in each transformer block. We find that \emph{tuning the vanilla linear attention module yields a strong approximation to the original softmax attention}, outperforming other linear attention variants~(Fig. \ref{fig:lolcats_vs_ours_stage1} and \ref{fig:all_attn_losses_stage1}). Although the linear attention modules are aligned independently in each block during Stage 1, the residual approximation error accumulates across layers. To mitigate this accumulated error, we introduce a feature alignment stage that finetunes the entire linearized model. Specifically, we replace the original softmax attention with the aligned linear attention in Stage 1, and finetune the full linearized ViT to align its final-layer features with those of the frozen softmax-based teacher model. Interestingly, we observe that \emph{the attention alignment in Stage 1 can accelerate convergence during this feature alignment process}. Finally, we perform supervised fine-tuning to transfer the adapted prior knowledge to downstream tasks. Our contributions are three-fold:

\begin{itemize}[leftmargin=*]
    \item We introduce a new paradigm for ViTs with linear attention that shifts the focus from designing more accurate attention approximations to adapting prior knowledge from pretrained ViTs. Our paradigm enables linearized ViTs to inherit the power of existing VFMs, eliminating the need for expensive training from scratch.
    \item We introduce \emph{ViT-AdaLA}, which adapts VFMs via attention alignment, feature alignment, and supervised fine-tuning. This progressive alignment allows linear attention models to inherit the strong priors of softmax-based ViTs. Furthermore, our framework is architecture-agnostic and compatible with other linear attention methods.
    \item We perform extensive experiments on classification and segmentation tasks across multiple VFMs, and compare against a wide range of state-of-the-art linear attention baselines.  Experimental results validate the effectiveness, efficiency, and scalability in resolution of ViT-AdaLA across different VFMs and downstream tasks.
\end{itemize}

\section{Related Work}
\textbf{Efficient Attention}. The Transformer architecture \cite{vaswani2017attention} has been widely adopted in both natural language processing and  vision tasks due to its  scalability. However, the quadratic complexity of standard attention limits long-context understanding, leading to numerous approaches to reduce \textit{memory} and \textit{computation} overhead.

FlashAttention \cite{dao2022flashattention, dao2023flashattention, shah2024flashattention} improves memory efficiency by employing tile-based computation instead of explicitly materializing the full attention matrix. To further reduce the number of visual tokens and improve computational efficiency, some methods either select informative tokens \cite{rao2021dynamicvit} or merge redundant ones \cite{bolya2023token,zeng2022not,li2025window}. Others propose to distill knowledge from a large ViT to a smaller one \cite{xiong2024efficientsam, touvron2021training} or a more efficient model \cite{bickllamba,wei2025vit}. Swin Transformer \cite{liu2021swin,liu2022swin} introduces a shifted-window mechanism to restrict dense attention computation within local regions. More recently, Mamba-based architectures \cite{gu2023mamba, liu2024vmamba, zhu2024vision, wang2025adventurer} have drawn significant attention due to their linear complexity, achieved through selective state-space modeling. 
Notably, Mamba can be seen as a variant of linear attention with specialized linear attention and modified block design \cite{han2024demystify}.

\textbf{Linear Attention}. Existing linearized Transformers can be broadly categorized into two streams: training-from-scratch-based and linearization-based approaches. Training-from-scratch-based approach targets at designing accurate attention  approximation methods and trains from scratch to obtain prior knowledge. One stream designs alternative activation functions after queries and keys for better approximation~\cite{han2024bridging,katharopoulos2020transformers,han2023flatten,shen2021efficient,qincosformer,bolya2022hydra,koohpayegani2024sima,ahmed2025mixa,bolya2022hydra}. Another family of methods employs low-rank decomposition to approximate, treating the softmax operation over queries and keys as a whole and decomposing it to derive more effective feature maps~\cite{xiong2021nystromformer,han2022modify,wu2024cur,yaras2025monarchattention, xu2024qt}. Recent work \cite{fan2025breaking} observes that rank augmentation is beneficial for improving the performance. Another stream tries to combine convolution kernel with linear attention, preserving local and global information \cite{zhou2025care, cai2023efficientvit}. However, these methods typically require large-scale pretraining before fine-tuning on downstream tasks, which is computationally and resource intensive. 

In contrast, linearization-based approaches aim to adapt existing softmax-based Transformers to linearized one. Hedgehog \cite{zhanghedgehog} approximates the attention matrix using the Hedgehog linear-attention module. LoLCATS~\cite{zhanglolcats} introduces attention transfer to approximate attention outputs and employs low-rank linearization based on LoRA \cite{hu2022lora} for decoder-based LLMs. Building upon LoLCATS, Lizard \cite{van2025lizard}, a hybrid attention paradigm, combines global attention via GLA \cite{yang2024gated} with local attention. Nevertheless, these methods cannot be directly applied to vision tasks due to architectural differences, as illustrated in Fig.~\ref{fig:enc-dec-comparison}. To address this challenge, we propose \textbf{ViT-AdaLA}, a novel method to extend the linearization paradigm to ViTs.
\vspace{-5pt}
\section{Method}
\subsection{Preliminary}
First, we briefly review the fundamentals of softmax and linear attention.
\begin{figure}[t]
    \centering
    \includegraphics[width=1\linewidth]{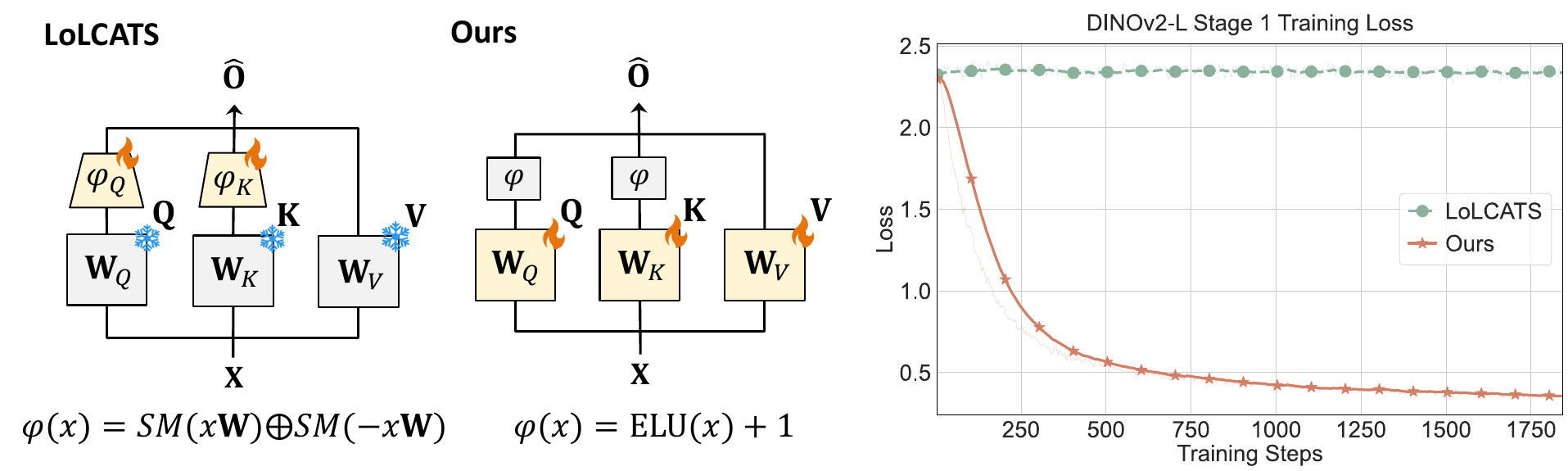}
    \caption{Linear attention architecture and Stage 1 training loss comparison with LoLCATS ($SM$: softmax; $\oplus$: concatenation). LoLCATS approximates the attention output based on Hedgehog \cite{zhanghedgehog} by tuning only two additional mapping modules applied to the queries and keys individually. In contrast, we tune all query, key, and value weights to approximate the attention output, which is both more efficient and more effective than the original LoLCATS approach.}
    \label{fig:lolcats_vs_ours_stage1}
\end{figure}
\begin{figure*}[t]
    \centering
    \includegraphics[width=1\linewidth]{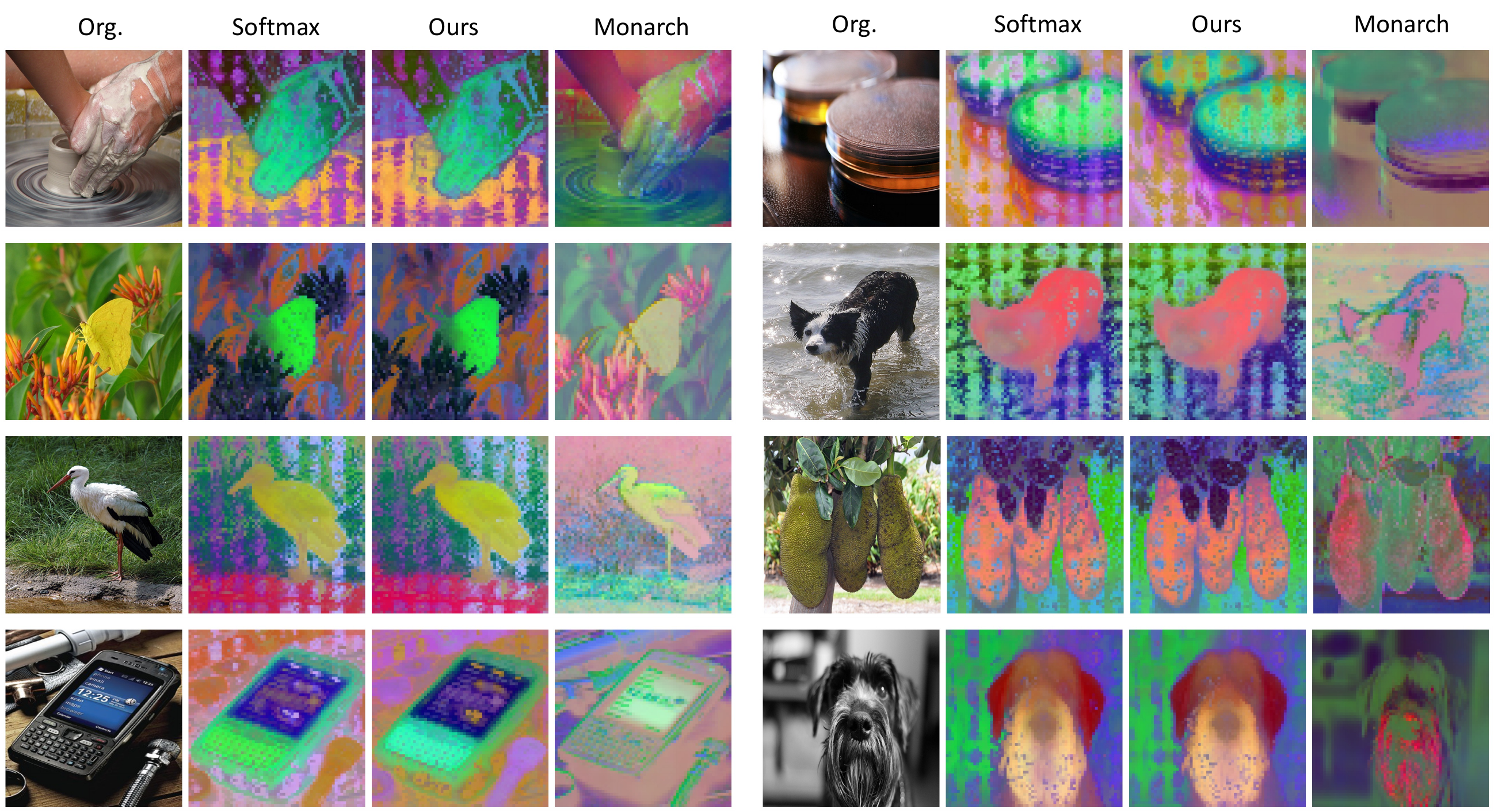}
    \caption{Visualization of PCA-projected features from the final layer of DINOv2-L. We compare the softmax-based features with those produced by ViT-AdaLA and Monarch attention \cite{yaras2025monarchattention} by projecting them to three channels using PCA. The results indicate that ViT-AdaLA better preserves the prior feature knowledge of the VFM. We provide more visualization results in the Appendix \ref{sec:vis_results}.}
    \label{fig:PCA_visualization}
\end{figure*}

\textbf{Softmax attention}. Softmax attention is the fundamental module of the original Transformer, responsible for computing the pairwise attention among all input tokens. Let $\mathbf{X}\in \mathbb{R}^{N\times D}$ denote a sequence of $N$ tokens, each with dimension $D$. The output $\mathbf{O}\in \mathbb{R}^{N\times D}$ is given by
\begin{equation}
\mathbf{O}_i=\sum_{j=1}^N{\frac{\exp(\mathbf{Q}_i \mathbf{K}_j^T)}{\sum_{k=1}^N \exp(\mathbf{Q}_i \mathbf{K}_k^T)} \mathbf{V}_j},
\end{equation}
where $\mathbf{Q}$, $\mathbf{K}$, and $\mathbf{V}$ denote the query, key, and value representations of the input tokens, obtained by multiplying $\mathbf{X}$ with the corresponding projection matrices $\mathbf{W}_Q$, $\mathbf{W}_K$, and $\mathbf{W}_V$, respectively. Here, $\exp(\cdot)$ denotes the exponential function, and we omit the common scaling factor $1/\sqrt{D}$ for simplicity. The computational complexity of softmax attention is $\mathcal{O}(N^2D)$.

\textbf{Linear Attention}. The kernel trick, expressed as $\exp(\mathbf{Q}\mathbf{K}^T) = \phi(\mathbf{Q})\phi(\mathbf{K}^T)$, is employed to decompose the multiplication of $\mathbf{Q}$ and $\mathbf{K}$ and to reorder the computation:
\begin{equation}\label{eq:linear_attention}
    \begin{split}        
    \mathbf{O}_i&=\sum_{j=1}^N{\frac{\phi(\mathbf{Q}_i) \phi(\mathbf{K}_j^T)}{\sum_{k=1}^N \phi(\mathbf{Q}_i) \phi(\mathbf{K}_k^T)} \mathbf{V}_j} =\frac{\phi(\mathbf{Q}_i) \sum_{j=1}^N{\phi(\mathbf{K}_j^T)\mathbf{V}_j}}{\phi(\mathbf{Q}_i) \sum_{k=1}^N \phi(\mathbf{K}_k^T)},
    \end{split}
\end{equation}
where $\phi(\cdot)=\mathrm{ELU}(\cdot)+1$, $\mathrm{ELU}(\cdot)$ indicates the exponential linear unit \cite{clevert2015fast}. We also compare with other types in the Appendix \ref{sec:activation_choices}. By reordering the multiplication computation of $\mathbf{Q}$ and $\mathbf{K}$, linear attention can achieve a computational complexity of $\mathcal{O}(ND^2)$.
\subsection{ViT-AdaLA}

ViT-AdaLA consists of three stages (see Fig.~\ref{fig:ViT-AdaLA}): attention alignment, feature alignment, and supervised fine-tuning.

\textbf{Stage 1: Attention Alignment}. To preserve the original attention quality while approximating softmax attention, we introduce an additional linear attention module and align it with the corresponding softmax attention module. 
Rather than adapting the linear attention module from scratch, we adapt it from an existing softmax attention module (based on Eq.~\ref{eq:linear_attention}) by simply modifying the computation order of queries, keys, and values using kernel trick. All components of the original model are frozen, except for the added linear attention module, where we only update the three projection matrices $\mathbf{W}_Q, \mathbf{W}_K$, and $\mathbf{W}_V$.

Formally, let the input to the $i$-th block after the first layer normalization be denoted as $\mathbf{O}_i\in \mathbb{R}^{N\times D}$. The output of the original self-attention module is ${\mathbf{{O}}}_i=f^{SA}_{\theta}(\mathbf{X}_i)\in \mathbb{R}^{N\times D}$, where $f_\theta^{SA}(\cdot)$ denotes the softmax-based self-attention. The attention alignment loss $\mathcal{L}_{att}$ is then defined as:
\begin{equation} \mathcal{L}_{att} = \frac{1}{N \cdot D} \sum_{i=1}^{K} \sum_{n=1}^{N} \sum_{d=1}^{D} \left( \mathbf{O}_i^{nd} - \hat{\mathbf{O}}_i^{nd} \right)^2, 
\end{equation}
where $\hat{\mathbf{O}}_i=f_\theta^{LA}(\mathbf{X}_i)$ denotes the output of the linear attention module, $n$ and $d$ index the token and feature dimension, respectively, and $K$ is the number of layers. The alignment loss is defined using the mean-square-error (MSE), which measures the discrepancy between the feature maps produced by the self-attention and linear-attention modules in each block. Importantly, the original features remain unchanged. We only adjust the linear-attention module to better approximate the behavior of the softmax self-attention.

Unlike the attention transfer strategy in LoLCATS \cite{zhanglolcats}, which tunes only two additional mapping modules applied to the queries and keys (\textit{i.e.}, Hedgehog linear attention), we adopt a vanilla linear attention formulation that relies on a simple activation function and directly tunes the query, key, and value projection matrices. We posit that the vanilla linear attention is highly malleable. Unlike sophisticated approximations whose rigid structural priors can ``fight'' the teacher during distillation, the vanilla attention’s unconstrained nature avoids optimization bottlenecks, allowing it to flexibly learn necessary approximation patterns. As shown in Fig.~\ref{fig:lolcats_vs_ours_stage1}, this design offers two key advantages compared to Hedgehog-based methods: (i) higher computational efficiency, and (ii) improved approximation quality.

\textbf{Stage 2: Feature Alignment}. Although we align the linear attention module with the softmax-based self-attention in each transformer block, the original features remain untuned, and replacing self-attention with linear attention introduces residual approximation errors to accumulate across blocks (see Fig. \ref{fig:pca_ablation} in Appendix \ref{sec:vis_results}). To ensure that the final output features of the linearized ViT remain consistent with the original model, we directly align the final features of the two models. Benefiting from the attention alignment in Stage 1, the linearized ViT converges faster and more effectively transfers prior knowledge from VFMs (see Sec. \ref{sec:stage_ablation}).

Specifically, we replace all softmax-based self-attention modules with the linear-attention modules obtained in Stage 1, resulting in a linearized ViT. This linearized ViT is then aligned with the frozen original ViT. Given the same input image $\mathbf{X}_v\in \mathbb{R}^{N\times D}$ for both models, we define the feature alignment loss $\mathcal{L}_{fa}$ as follows:\vspace{-5pt}
\begin{equation}
    \mathcal{L}_{fa} = \frac{\lambda}{N\cdot D}\sum_{n=1}^{N}\sum_{d=1}^{D}(\mathbf{f}_v^{nd}-\hat{\mathbf{f}}_v^{nd})^2,
\end{equation}
where $\mathbf{f}_v=\mathcal{F}_{\theta_0}(\mathbf{X}_v)$ and $\hat{\mathbf{f}}_v=\mathcal{F}_{\theta}(\mathbf{X}_v)$ denote the final representations produced by the original ViT and the linearized ViT, respectively. $\lambda$ controls the scale of the output loss and is set to different values for different VFMs.  We utilize MSE loss to align two final features. During this stage, $\mathcal{F}_{\theta_0}$ is kept frozen, while only $\mathcal{F}_{\theta}$ is updated.

\textbf{Stage 3: Supervised Fine-tuning}. After feature alignment, we transfer the linearized ViT, enriched with prior knowledge from VFMs, to downstream tasks by fine-tuning it on task-specific datasets. In this stage, a task-specific head is appended to the linearized ViT and both the backbone and the task head are updated during this process.

\begin{table*}[t]
\centering
\caption{Top-1 fine-tuning accuracy comparison on ImageNet-1K under different vision foundation models with multiple linear attention baselines. We reproduce all baselines to ensure a fair comparison. The classification head is a single linear layer for all the methods. }
\label{tab:imagenet_benchmarks}
\setlength{\tabcolsep}{9pt}
\renewcommand{\arraystretch}{1.1}
\scalebox{0.7}{
\begin{tabular}{p{3.7cm} c c c c c c c}
\toprule
& \multirow{2}{*}{\textbf{Backbone}} & \multirow{2}{*}{\textbf{Res.}} & \textbf{Params} & {\textbf{FLOPS}} & \textbf{Peak Mem.} & {\textbf{Throughput}} & \textbf{Top-1 Acc.} \\
& & & \textbf{(M)} &\textbf{(G)} & \textbf{(GB)} & \textbf{(imgs/s)} & \textbf{(\%)} \\
\midrule
\rowcolor{gray!10} Softmax \cite{oquab2024dinov2} & DINOv2-L & $512^2$ & 304.20 & 310.60 & 1.3181 & 36.52 & 86.8 \\
Hedgehog \cite{zhanghedgehog}  & DINOv2-L & $512^2$ & 305.77 & 265.41\textcolor{darkgreen}{$_{\downarrow 14.5\%}$} & 1.2221\textcolor{darkgreen}{$_{\downarrow 7.3\%}$} & 37.44\textcolor{darkgreen}{$_{\uparrow 2.5\%}$}  & 58.8 \\
LoLCATS \cite{zhanglolcats}  & DINOv2-L & $512^2$ & 305.77 & 265.41\textcolor{darkgreen}{$_{\downarrow 14.5\%}$} & 1.2221\textcolor{darkgreen}{$_{\downarrow 7.3\%}$} & 37.44\textcolor{darkgreen}{$_{\uparrow 2.5\%}$} & 61.6 \\
Linformer \cite{wang2020linformer} & DINOv2-L & $512^2$ & 305.77 & 262.19\textcolor{darkgreen}{$_{\downarrow 15.6\%}$} & 1.2221\textcolor{darkgreen}{$_{\downarrow 7.3\%}$} & 45.41\textcolor{darkgreen}{$_{\uparrow 25.3\%}$} & 68.9 \\
Performer \cite{choromanskirethinking} & DINOv2-L & $512^2$ & 304.20 & 265.41\textcolor{darkgreen}{$_{\downarrow 14.6\%}$} & 1.2164\textcolor{darkgreen}{$_{\downarrow 7.7\%}$} & 36.48\textcolor{darkgreen}{$_{\uparrow 0.1\%}$} & 75.2 \\
Cosformer \cite{qincosformer} & DINOv2-L & $512^2$ & 304.20 & 265.41\textcolor{darkgreen}{$_{\downarrow 14.6\%}$} & 1.2226\textcolor{darkgreen}{$_{\downarrow 7.2\%}$} & 39.55\textcolor{darkgreen}{$_{\uparrow 8.2\%}$} & 75.1 \\
Nystr\"{o}mformer  \cite{xiong2021nystromformer} & DINOv2-L & $512^2$ & 304.20 & 265.20\textcolor{darkgreen}{$_{\downarrow 14.6\%}$} & 1.2163\textcolor{darkgreen}{$_{\downarrow 7.7\%}$} & 18.29\textcolor{darkred}{$_{\downarrow 50.8\%}$} & 82.4 \\
Monarch \cite{yaras2025monarchattention} & DINOv2-L & $512^2$ & 304.20 & 269.44\textcolor{darkgreen}{$_{\downarrow 13.3\%}$} & 1.2304\textcolor{darkgreen}{$_{\downarrow 6.7\%}$} & 18.31\textcolor{darkred}{$_{\downarrow 48.9\%}$} & 82.7 \\
 \rowcolor{lightyellow} ViT-AdaLA (Stage 2) & DINOv2-L & $512^2$ & 304.20 & 262.19\textcolor{darkgreen}{$_{\downarrow 15.6\%}$} & 1.2163\textcolor{darkgreen}{$_{\downarrow 7.7\%}$} & 41.56\textcolor{darkgreen}{$_{\uparrow 16.1\%}$} & 84.5 \\
\rowcolor{lightyellow} ViT-AdaLA (Ours) & DINOv2-L & $512^2$ & 304.20 & 262.19\textcolor{darkgreen}{$_{\downarrow 15.6\%}$} & 1.2163\textcolor{darkgreen}{$_{\downarrow 7.7\%}$} & 41.56\textcolor{darkgreen}{$_{\uparrow 16.1\%}$} & 86.0 \\
\midrule
\rowcolor{gray!10} Softmax \cite{radford2021learning} & CLIP-L & $512^2$ & 304.15 & 310.60 & 1.3179 & 36.82 & 86.4 \\
\rowcolor{lightyellow} ViT-AdaLA (Stage 2) & CLIP-L & $512^2$ & 304.15 & 262.19\textcolor{darkgreen}{$_{\downarrow 15.6\%}$}  & 1.2161\textcolor{darkgreen}{$_{\downarrow 7.4\%}$} & 42.14\textcolor{darkgreen}{$_{\uparrow 15.7\%}$} & 83.4 \\
\rowcolor{lightyellow} ViT-AdaLA (Ours) & CLIP-L & $512^2$ & 304.15 & 262.19\textcolor{darkgreen}{$_{\downarrow 15.6\%}$}  & 1.2161\textcolor{darkgreen}{$_{\downarrow 7.4\%}$} & 42.14\textcolor{darkgreen}{$_{\uparrow 15.7\%}$} & 85.5 \\
\midrule
\rowcolor{gray!10} Softmax \cite{zhai2023sigmoid} & SigLIP-L & $512^2$ & 316.74 & 312.46 & 1.3636 & 40.86 & 86.9 \\
\rowcolor{lightyellow} ViT-AdaLA (Stage 2) & SigLIP-L & $512^2$ & 316.74 & 264.04\textcolor{darkgreen}{$_{\downarrow 15.5\%}$} & 1.2620\textcolor{darkgreen}{$_{\downarrow 7.4\%}$} & 46.91\textcolor{darkgreen}{$_{\uparrow 13.9\%}$} & 84.9 \\
\rowcolor{lightyellow} ViT-AdaLA (Ours) & SigLIP-L & $512^2$ & 316.74 & 264.04\textcolor{darkgreen}{$_{\downarrow 15.5\%}$} & 1.2620\textcolor{darkgreen}{$_{\downarrow 7.4\%}$} & 46.91\textcolor{darkgreen}{$_{\uparrow 13.9\%}$} & 86.4 \\
\bottomrule
\end{tabular}
}
\end{table*}
\section{Experiment}

We pretrain linearized VFMs using our ViT-AdaLA pipeline through Stages 1 and 2. Specifically, we train four linearized VFMs within the PyTorch Lightning framework using 8 $\times$ H100 GPUs. Stage 1 training is conducted on COCO \cite{lin2014microsoft} for 4 epochs with batch size 32 per GPU, while Stage 2 training is performed on ImageNet-22K \cite{deng2009imagenet} for 10 to 30 epochs with batch size 16 per GPU. We employ the AdamW optimizer with fixed learning rate $1e^{-2}$ and initial learning rate $1e^{-4}$ and a linearly decaying learning rate schedule for Stage 2. We multiply the learning rate with the ratio of 0.1 to the backbone when training. All models are trained using $512\times512$ input images, with random cropping and color jitter applied for data augmentation. More details can refer to Appendix \ref{sec:hyper-parameter_configs}.

After pretraining, we benchmark performance on classification and semantic segmentation against existing linear attention baselines. Additionally, we perform ablation studies to analyze the impact of each training stage.
\begin{table*}[t]
\centering
\caption{mIoU fine-tuning comparison on ADE20K under different vision foundation models with multiple linear attention baselines. We reproduce all baselines to ensure a fair comparison. The segmentation head is Mask2former for all the methods.}
\label{tab:ade20k_bench}
\setlength{\tabcolsep}{10pt}
\renewcommand{\arraystretch}{1.1}
\scalebox{0.7}{
\begin{tabular}{p{3.5cm} c c c c c c c}
\toprule
& \multirow{2}{*}{\textbf{Backbone}} & \multirow{2}{*}{\textbf{Res.}} & \textbf{Params} & {\textbf{FLOPS}} & \textbf{Peak Mem.} & {\textbf{Throughput}} & \textbf{mIoU} \\
& & & \textbf{(M)} & \textbf{(G)} & \textbf{(GB)} & \textbf{(imgs/s)} & \textbf{(\%)} \\
\midrule
\rowcolor{gray!10} Softmax \cite{oquab2024dinov2} & DINOv2-L & $512^2$ & 304.20 & 310.60 & 1.3181 & 36.52 & 56.73 \\
Hedgehog \cite{zhanghedgehog}  & DINOv2-L & $512^2$ & 305.77 & 265.41\textcolor{darkgreen}{$_{\downarrow 14.5\%}$} & 1.2221\textcolor{darkgreen}{$_{\downarrow 7.3\%}$} & 37.44\textcolor{darkgreen}{$_{\uparrow 2.5\%}$}  & 22.52 \\
LoLCATS \cite{zhanglolcats}  & DINOv2-L & $512^2$ & 305.77 & 265.41\textcolor{darkgreen}{$_{\downarrow 14.5\%}$} & 1.2221\textcolor{darkgreen}{$_{\downarrow 7.3\%}$} & 37.44\textcolor{darkgreen}{$_{\uparrow 2.5\%}$} & 17.42 \\
Linformer \cite{wang2020linformer} & DINOv2-L & $512^2$ & 305.77 & 262.19\textcolor{darkgreen}{$_{\downarrow 15.6\%}$} & 1.2221\textcolor{darkgreen}{$_{\downarrow 7.3\%}$} & 45.41\textcolor{darkgreen}{$_{\uparrow 25.3\%}$} & 18.82 \\
Performer \cite{choromanskirethinking} & DINOv2-L & $512^2$ & 304.20 & 265.41\textcolor{darkgreen}{$_{\downarrow 14.6\%}$} & 1.2164\textcolor{darkgreen}{$_{\downarrow 7.7\%}$} & 36.48\textcolor{darkgreen}{$_{\uparrow 0.1\%}$} & 41.16 \\
Cosformer \cite{qincosformer} & DINOv2-L & $512^2$ & 304.20 & 265.41\textcolor{darkgreen}{$_{\downarrow 14.6\%}$} & 1.2226\textcolor{darkgreen}{$_{\downarrow 7.2\%}$} & 39.55\textcolor{darkgreen}{$_{\uparrow 8.2\%}$} & 27.17 \\
Nystr\"{o}mformer  \cite{xiong2021nystromformer} & DINOv2-L & $512^2$ & 304.20 & 265.20\textcolor{darkgreen}{$_{\downarrow 14.6\%}$} & 1.2163\textcolor{darkgreen}{$_{\downarrow 7.7\%}$} & 18.29\textcolor{darkred}{$_{\downarrow 50.8\%}$} & 41.16 \\
Monarch \cite{yaras2025monarchattention} & DINOv2-L & $512^2$ & 304.20 & 269.44\textcolor{darkgreen}{$_{\downarrow 13.3\%}$} & 1.2304\textcolor{darkgreen}{$_{\downarrow 6.7\%}$} & 18.31\textcolor{darkred}{$_{\downarrow 48.9\%}$} & 44.95 \\
 \rowcolor{lightyellow} ViT-AdaLA (Stage 2) & DINOv2-L & $512^2$ & 304.20 & 262.19\textcolor{darkgreen}{$_{\downarrow 15.6\%}$} & 1.2163\textcolor{darkgreen}{$_{\downarrow 7.7\%}$} & 41.56\textcolor{darkgreen}{$_{\uparrow 16.1\%}$} & 52.46 \\
\rowcolor{lightyellow} ViT-AdaLA (Ours) & DINOv2-L & $512^2$ & 304.20 & 262.19\textcolor{darkgreen}{$_{\downarrow 15.6\%}$} & 1.2163\textcolor{darkgreen}{$_{\downarrow 7.7\%}$} & 41.56\textcolor{darkgreen}{$_{\uparrow 16.1\%}$} & 55.55 \\
\midrule
\rowcolor{gray!10} Softmax \cite{zhai2023sigmoid} & SigLIP-L & $512^2$ & 316.74 & 312.46 & 1.3636 & 41.05 & 54.4 \\
\rowcolor{lightyellow} ViT-AdaLA (Stage 2) & SigLIP-L & $512^2$ & 316.74 & 264.04\textcolor{darkgreen}{$_{\downarrow 15.5\%}$} & 1.2620\textcolor{darkgreen}{$_{\downarrow 7.4\%}$} & 47.05\textcolor{darkgreen}{$_{\uparrow 14.6\%}$} & 51.39 \\
\rowcolor{lightyellow} ViT-AdaLA (Ours) & SigLIP-L & $512^2$ & 316.74 & 264.04\textcolor{darkgreen}{$_{\downarrow 15.5\%}$} & 1.2620\textcolor{darkgreen}{$_{\downarrow 7.4\%}$} & 47.05\textcolor{darkgreen}{$_{\uparrow 14.6\%}$} & 53.16 \\
\midrule
\rowcolor{gray!10} Softmax & IN1K ViT-L  & $512^2$ & 304.15 & 310.60 & 1.3179 & 36.47 & 50.83 \\
\rowcolor{lightyellow} ViT-AdaLA ({\text{Stage 2}}) & IN1K ViT-L   & $512^2$ & 304.15 & 262.19\textcolor{darkgreen}{$_{\downarrow 15.6\%}$} & 1.2161\textcolor{darkgreen}{$_{\downarrow 7.7\%}$} & 42.56\textcolor{darkgreen}{$_{\uparrow 16.7\%}$} & 46.92 \\
\rowcolor{lightyellow} ViT-AdaLA (Ours) & IN1K ViT-L   & $512^2$ & 304.15 & 262.19\textcolor{darkgreen}{$_{\downarrow 15.6\%}$} & 1.2161\textcolor{darkgreen}{$_{\downarrow 7.7\%}$} & 42.56\textcolor{darkgreen}{$_{\uparrow 16.7\%}$} & 49.75 \\
\bottomrule
\end{tabular}
}
\end{table*}

\subsection{Comparison on Classification}
\textbf{Experimental setup.} We conduct experiments on the ImageNet-1K \cite{deng2009imagenet} dataset. We report top-1 accuracy, parameter, throughput, and GFLOPs in Tab.~\ref{tab:imagenet_benchmarks}. Throughput and peak memory (batch size 1) are measured on a single H100 GPU. This measurement setup extends to Tables \ref{tab:ade20k_bench} and \ref{tab:cityscapes_bench}. The baselines are constructed by replacing softmax attention modules with their linear counterparts. Full training details are provided in the Appendix \ref{sec:hyper-parameter_configs}.

\begin{wrapfigure}{r}{0.48\linewidth}
    \centering
    \vspace{-15pt}
    \includegraphics[width=\linewidth]{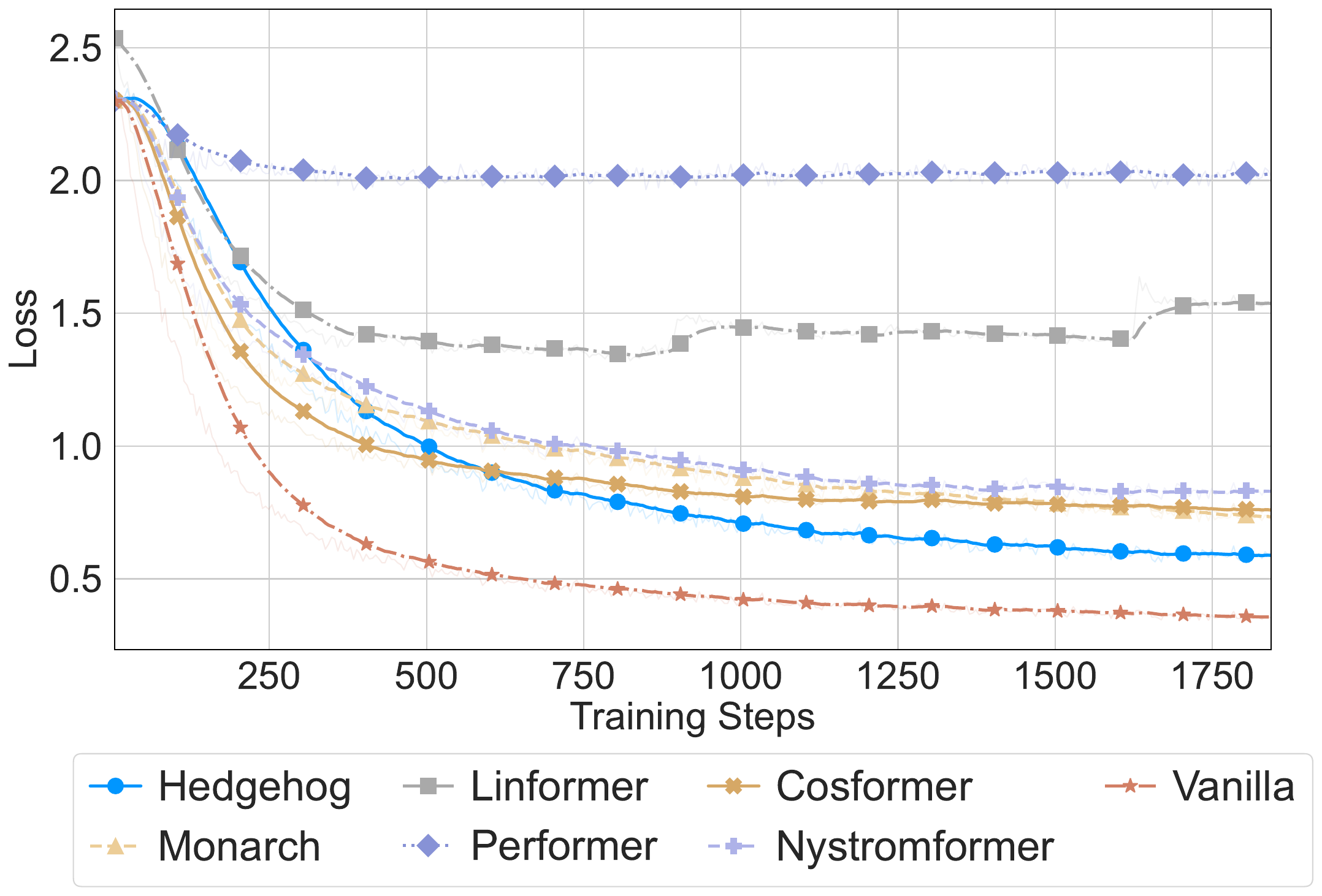}
    \caption{Training loss comparison of different linear attention variants in Stage 1 on DINOv2-L. To ensure a fair comparison, the query, key, and value weights are tuned in every layer for each baseline. Vanilla linear attention exhibits superior approximation performance compared to the other variants.}
    \label{fig:all_attn_losses_stage1}
    \vspace{-65pt}
\end{wrapfigure}
\textbf{Result analysis.} Tab. \ref{tab:imagenet_benchmarks} shows that ViT-AdaLA achieves the highest top-1 accuracy among VFMs, \emph{maintaining accuracy within 1\% of the original softmax backbone while preserving efficiency}. We can also observe that for decoder-based linearization methods like Hedgehog \cite{zhanghedgehog} and LoLCATS \cite{zhanglolcats}, the final performance drops significantly since the linearized ViT has not been fully aligned with the VFM backbone, and \emph{aligning attention alone is insufficient to transfer adequate prior knowledge}. This demonstrates that:
\vspace{-2pt}
\begin{tcolorbox}[
    enhanced,
    colback=gray!5,          
    colframe=teal!30!black,  
    left=3mm,right=3mm,
    top=1mm,bottom=1mm,
    boxrule=0.8pt,           
    arc=4pt,                 
    drop shadow              
]
\itshape
For vision models, it is essential to align the final feature representations with the original VFM backbone teacher to ensure effective knowledge transfer.
\end{tcolorbox}
For training-from-scratch methods, low-rank approximation yields better than activation-based techniques. However, they incur greater memory and computation due to the mathematical complexity required for high-quality approximation. Nevertheless, these methods still fail to match the performance of ViT-AdaLA or even its Stage 2 baseline. This indicates: \emph{linearization is superior to training-from-scratch methods for extracting prior knowledge from VFMs.
}
\subsection{Comparison on Semantic Segmentation}

\textbf{Experimental setup.} We further conduct experiments on ADE20K \cite{zhou2017scene} and Cityscapes \cite{cordts2016cityscapes} to provide a more fine-grained evaluation of ViT-AdaLA when transferring from VFMs. For both semantic segmentation datasets, we employ the Mask2Former head~\cite{cheng2022masked} across all baselines. We consider two experimental settings: evaluating different VFMs on ADE20K in Tab. \ref{tab:ade20k_bench}, and assessing the impact of input resolutions on Cityscapes in Tab. \ref{tab:cityscapes_bench}.

\textbf{Result analysis.} 
Since segmentation requires more low-level and fine-grained features than classification, the ability of linearization to extract robust prior knowledge is essential for maintaining high performance in dense prediction tasks. 
As shown in Tab. \ref{tab:ade20k_bench}, ViT-AdaLA demonstrates strong performance across various VFMs, rivaling even supervised baselines such as the IN1K-pretrained ViT \cite{dosovitskiyimage} 
This highlights \emph{the generalizability of ViT-AdaLA in effectively distilling prior knowledge from diverse VFMs and transferring it to different downstream tasks}.

We further explore the scaling ability of our ViT-AdaLA for higher resolution images as shown in Tab. \ref{tab:cityscapes_bench}. 
Our linear approach overcomes the efficiency bottleneck of softmax attention, delivering $>$50\% memory savings and 2$\times$ faster inference.

Moreover, ViT-AdaLA generalizes across scales: although distilled at $512^2$,  its performance improves from 72.40\% to 78.73\% when scaled up to $1024^2$, which shows the following property:
\begin{tcolorbox}[
    enhanced,
    colback=gray!5,          
    colframe=teal!30!black,  
    left=3mm,right=3mm,
    top=1mm,bottom=1mm,
    boxrule=0.8pt,           
    arc=4pt,                 
    drop shadow              
]
\itshape
ViT-AdaLA can scale to higher-resolution images, even pretrained on lower-resolution ones.
\end{tcolorbox}
Such a property enables more efficient pretraining and broader applications of ViT-AdaLA. Ultimately, this flexibility resolves the tension between training costs and inference quality, establishing ViT-AdaLA as a versatile and practical paradigm for large-resolution dense prediction tasks.

\begin{table*}[t]
\centering
\caption{mIoU fine-tuning comparison on Cityscapes under different input resolutions (512 vs. 1024) based on DINOv2-L. We reproduce all baselines to ensure a fair comparison. The segmentation head is Mask2former for all the methods.}
\label{tab:cityscapes_bench}
\setlength{\tabcolsep}{10pt}
\renewcommand{\arraystretch}{1.1}
\scalebox{0.7}{
\begin{tabular}{p{3.5cm} c c c c c c c}
\toprule
& \multirow{2}{*}{\textbf{Backbone}} & \multirow{2}{*}{\textbf{Res.}} & \textbf{Params} & {\textbf{FLOPS}} & \textbf{Peak Mem.} & {\textbf{Throughput}} & \textbf{mIoU} \\
& & & \textbf{(M)} & \textbf{(G)} & \textbf{(GB)} & \textbf{(imgs/s)} & \textbf{(\%)} \\
\midrule
\rowcolor{gray!10} Softmax \cite{oquab2024dinov2} & DINOv2-L & $512^2$ & 304.20 & 310.60 & 1.3181 & 36.52 & 74.86 \\
Hedgehog \cite{zhanghedgehog}  & DINOv2-L & $512^2$ & 305.77 & 265.41\textcolor{darkgreen}{$_{\downarrow 14.5\%}$} & 1.2221\textcolor{darkgreen}{$_{\downarrow 7.3\%}$} & 37.44\textcolor{darkgreen}{$_{\uparrow 2.5\%}$}  & 43.97 \\
LoLCATS \cite{zhanglolcats}  & DINOv2-L & $512^2$ & 305.77 & 265.41\textcolor{darkgreen}{$_{\downarrow 14.5\%}$} & 1.2221\textcolor{darkgreen}{$_{\downarrow 7.3\%}$} & 37.44\textcolor{darkgreen}{$_{\uparrow 2.5\%}$} & 34.71 \\
Linformer \cite{wang2020linformer} & DINOv2-L & $512^2$ & 305.77 & 262.19\textcolor{darkgreen}{$_{\downarrow 15.6\%}$} & 1.2221\textcolor{darkgreen}{$_{\downarrow 7.3\%}$} & 45.41\textcolor{darkgreen}{$_{\uparrow 25.3\%}$} & 37.24 \\
Performer \cite{choromanskirethinking} & DINOv2-L & $512^2$ & 304.20 & 265.41\textcolor{darkgreen}{$_{\downarrow 14.6\%}$} & 1.2164\textcolor{darkgreen}{$_{\downarrow 7.7\%}$} & 36.48\textcolor{darkgreen}{$_{\uparrow 0.1\%}$} & 43.11 \\
Cosformer \cite{qincosformer} & DINOv2-L & $512^2$ & 304.20 & 265.41\textcolor{darkgreen}{$_{\downarrow 14.6\%}$} & 1.2226\textcolor{darkgreen}{$_{\downarrow 7.2\%}$} & 39.55\textcolor{darkgreen}{$_{\uparrow 8.2\%}$} & 44.52 \\
Nystr\"{o}mformer  \cite{xiong2021nystromformer} & DINOv2-L & $512^2$ & 304.20 & 265.20\textcolor{darkgreen}{$_{\downarrow 14.6\%}$} & 1.2163\textcolor{darkgreen}{$_{\downarrow 7.7\%}$} & 18.29\textcolor{darkred}{$_{\downarrow 50.8\%}$} & 48.97 \\
Monarch \cite{yaras2025monarchattention} & DINOv2-L & $512^2$ & 304.20 & 269.44\textcolor{darkgreen}{$_{\downarrow 13.3\%}$} & 1.2304\textcolor{darkgreen}{$_{\downarrow 6.7\%}$} & 18.31\textcolor{darkred}{$_{\downarrow 48.9\%}$} & 50.84 \\
 \rowcolor{lightyellow} ViT-AdaLA (Stage 2) & DINOv2-L & $512^2$ & 304.20 & 262.19\textcolor{darkgreen}{$_{\downarrow 15.6\%}$} & 1.2163\textcolor{darkgreen}{$_{\downarrow 7.7\%}$} & 41.56\textcolor{darkgreen}{$_{\uparrow 16.1\%}$} & 71.32 \\
\rowcolor{lightyellow} ViT-AdaLA (Ours) & DINOv2-L & $512^2$ & 304.20 & 262.19\textcolor{darkgreen}{$_{\downarrow 15.6\%}$} & 1.2163\textcolor{darkgreen}{$_{\downarrow 7.7\%}$} & 41.56\textcolor{darkgreen}{$_{\uparrow 16.1\%}$} & 72.40 \\
\midrule
\rowcolor{gray!10} Softmax \cite{oquab2024dinov2} & DINOv2-L & $1024^2$ & 304.20 & 1241.0 & 3.2836 & 7.07 & 80.98 \\
Hedgehog \cite{zhanghedgehog}  & DINOv2-L & $1024^2$ & 305.77 & 442.24\textcolor{darkgreen}{$_{\downarrow 64.4\%}$} & 1.3823\textcolor{darkgreen}{$_{\downarrow 57.9\%}$} & 15.38\textcolor{darkgreen}{$_{\uparrow 117.4\%}$} & 44.17 \\
LoLCATS \cite{zhanglolcats}  & DINOv2-L & $1024^2$ & 305.77 & 442.24\textcolor{darkgreen}{$_{\downarrow 64.4\%}$} & 1.3823\textcolor{darkgreen}{$_{\downarrow 57.9\%}$} & 15.38\textcolor{darkgreen}{$_{\uparrow 117.4\%}$} & 33.66 \\
Linformer \cite{wang2020linformer} & DINOv2-L & $1024^2$ & 305.77 & 429.35\textcolor{darkgreen}{$_{\downarrow 65.4\%}$} & 1.3999\textcolor{darkgreen}{$_{\downarrow 57.4\%}$} & 17.83\textcolor{darkgreen}{$_{\uparrow 152.0\%}$} & 26.63 \\
Performer \cite{choromanskirethinking} & DINOv2-L & $1024^2$ & 304.20 & 442.24\textcolor{darkgreen}{$_{\downarrow 64.4\%}$} & 1.3765\textcolor{darkgreen}{$_{\downarrow 58.1\%}$} & 15.23\textcolor{darkgreen}{$_{\uparrow 115.3\%}$} & 42.58 \\
Cosformer \cite{qincosformer} & DINOv2-L & $1024^2$ & 304.20 & 442.24\textcolor{darkgreen}{$_{\downarrow 64.4\%}$} & 1.3945\textcolor{darkgreen}{$_{\downarrow 57.5\%}$} & 15.29\textcolor{darkgreen}{$_{\uparrow 116.1\%}$} & 44.50 \\
Nystr\"{o}mformer  \cite{xiong2021nystromformer} & DINOv2-L & $1024^2$ & 304.20 & 442.24\textcolor{darkgreen}{$_{\downarrow 64.4\%}$} & 1.3764\textcolor{darkgreen}{$_{\downarrow 58.1\%}$} & 14.61\textcolor{darkgreen}{$_{\uparrow 106.5\%}$} & 47.56 \\
Monarch \cite{yaras2025monarchattention} & DINOv2-L & $1024^2$ & 304.20 & 500.24\textcolor{darkgreen}{$_{\downarrow 59.7\%}$} & 1.5469\textcolor{darkgreen}{$_{\downarrow 52.9\%}$} & 9.74\textcolor{darkgreen}{$_{\uparrow 37.6\%}$} & 55.18 \\
\rowcolor{lightyellow} ViT-AdaLA (Stage 2) & DINOv2-L & $1024^2$ & 304.20 & 429.35\textcolor{darkgreen}{$_{\downarrow 65.4\%}$} & 1.3764\textcolor{darkgreen}{$_{\downarrow 58.1\%}$} & 15.95\textcolor{darkgreen}{$_{\uparrow 125.4\%}$} & 77.53 \\
\rowcolor{lightyellow} ViT-AdaLA (Ours) & DINOv2-L & $1024^2$ & 304.20 & 429.35\textcolor{darkgreen}{$_{\downarrow 65.4\%}$} & 1.3764\textcolor{darkgreen}{$_{\downarrow 58.1\%}$} & 15.95\textcolor{darkgreen}{$_{\uparrow 125.4\%}$} & 78.73 \\
\bottomrule
\end{tabular}
}
\end{table*}

\subsection{Ablation Study}


\subsubsection{Effectiveness of Pretraining Stages}\label{sec:stage_ablation}
We ablate Stages 1 and 2 on DINOv2-L/ADE20K and SAM-H/SAM-HQ (Table~\ref{tab:stage_ablation}). For SAM-H, training uses an internal foreground-segmentation dataset and evaluation uses SAM-HQ.

\begin{figure}[t]
    \centering

    \begin{minipage}[t]{0.36\linewidth}
        \centering
        \includegraphics[width=0.92\linewidth]{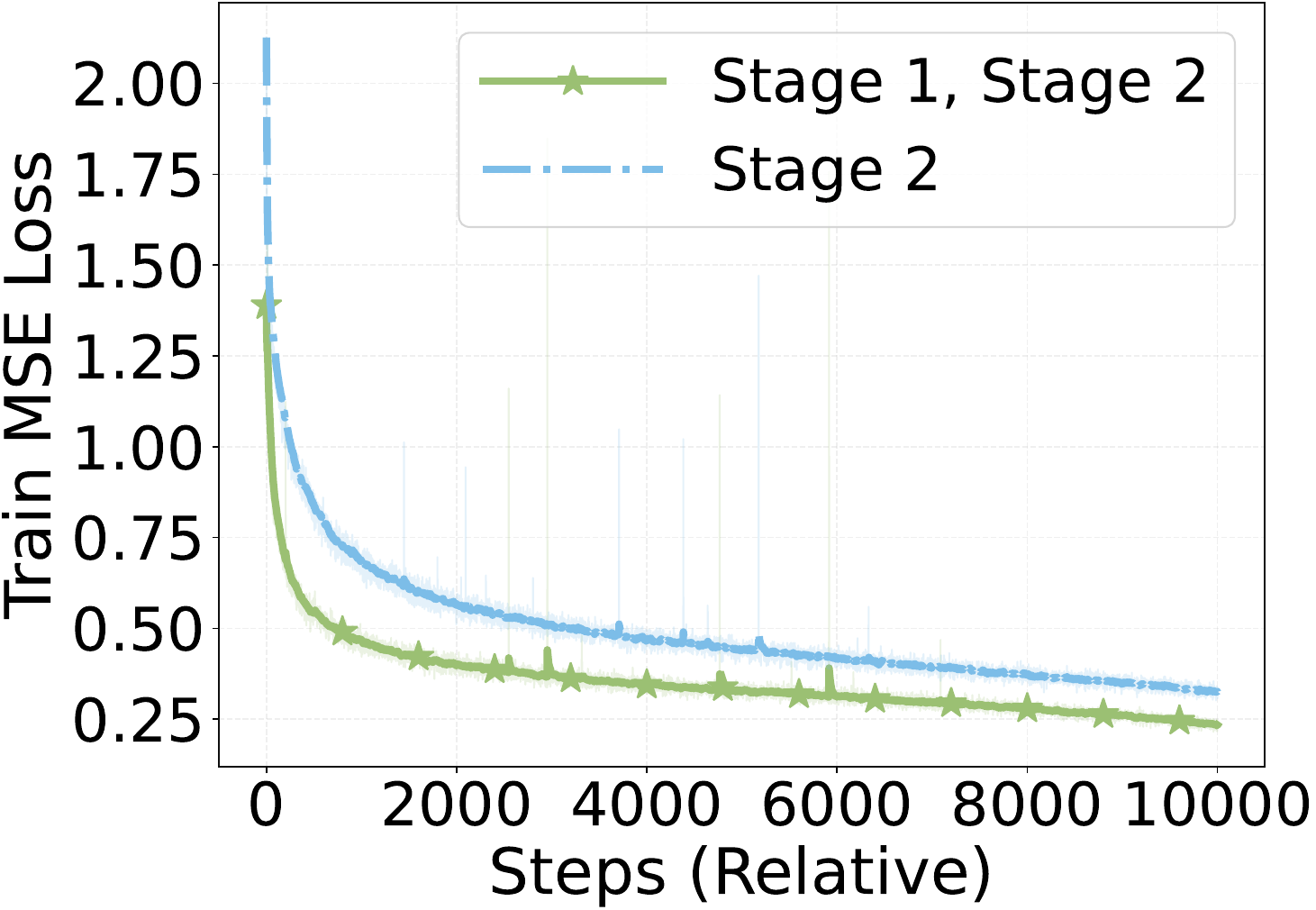}
        \captionof{figure}{Stage-2 feature-MSE loss with/without Stage 1 (DINOv2-L).}
        \label{fig:loss_comparison_w_wo_stage1}
    \end{minipage}
    \hfill
    \begin{minipage}[t]{0.31\linewidth}
        \centering
        \includegraphics[width=0.9\linewidth]{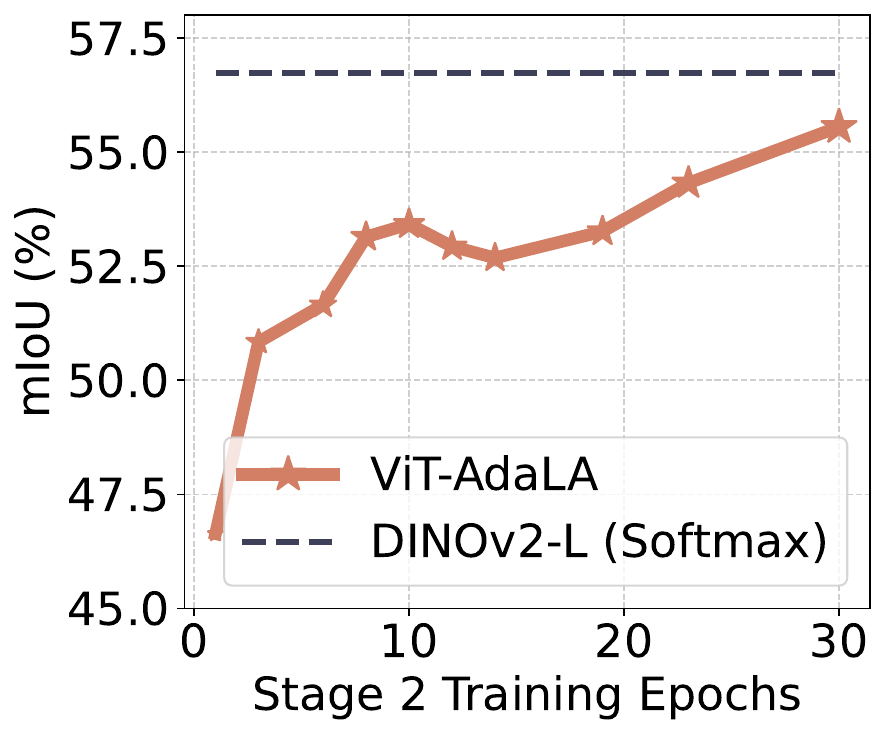}
        \captionof{figure}{ADE20K performance across different Stage 2 training epochs for DINOv2-L.}
        \label{fig:ade20k_performance_across_stage2_epochs}
    \end{minipage}
    \hfill
    \begin{minipage}[t]{0.3\linewidth}
        \centering
        \includegraphics[width=\linewidth]{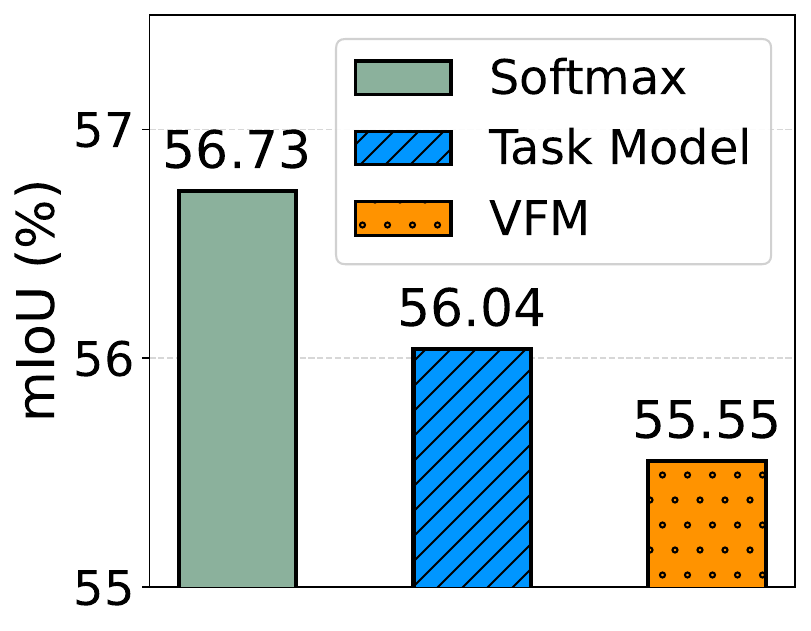}
        \captionof{figure}{ADE20K performance of adapting task model or VFM for DINOv2-L.
        }
        \label{fig:directly_adapting_task_model}
    \end{minipage}

\end{figure}

\begin{figure*}[t]
\vspace{-10pt}
\centering
\begin{minipage}[b]{0.32\linewidth}
\centering
\includegraphics[width=\linewidth]{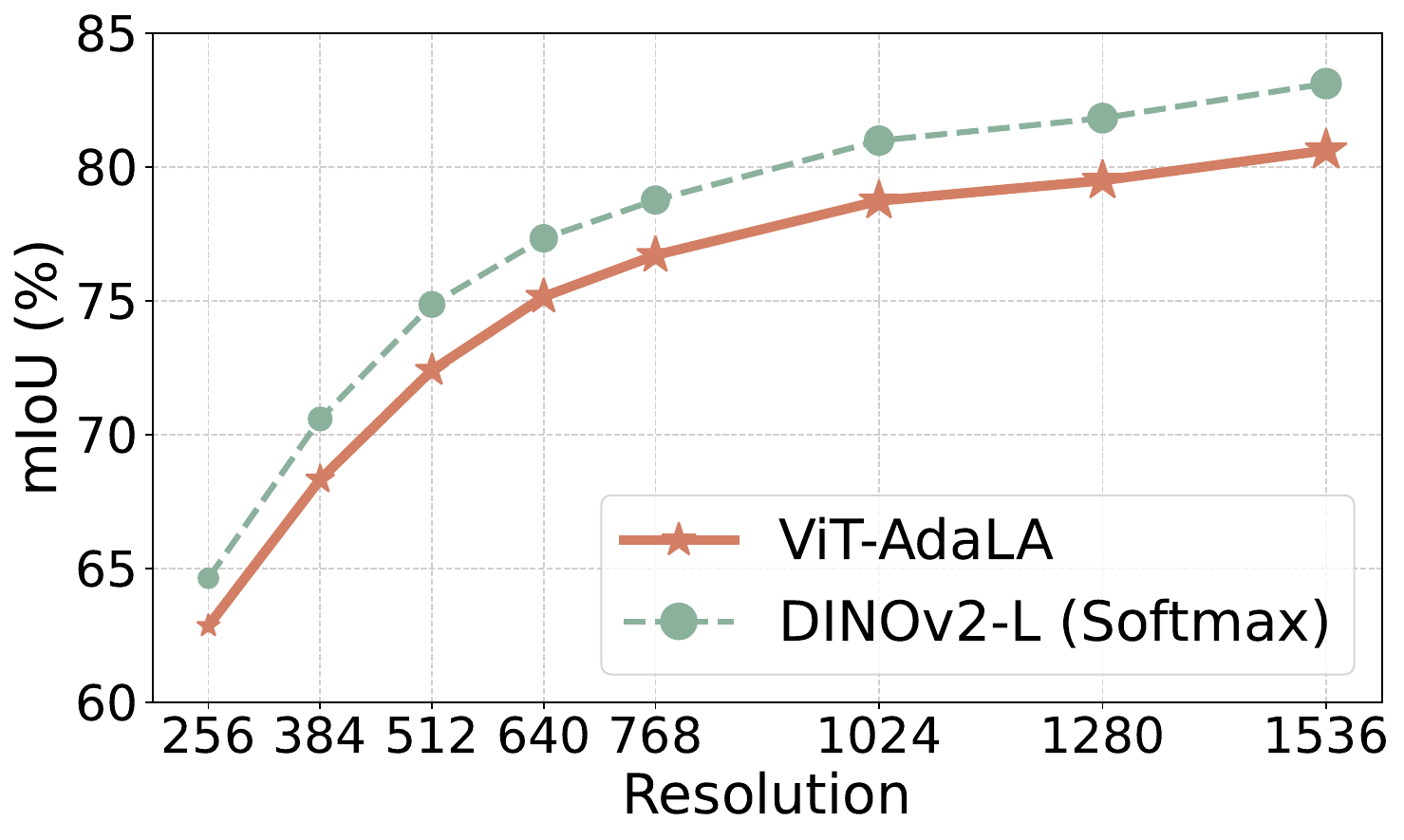}
\centerline{\small(a) mIoU}
\end{minipage}
\hfill
\begin{minipage}[b]{0.32\linewidth}
\centering
\includegraphics[width=\linewidth]{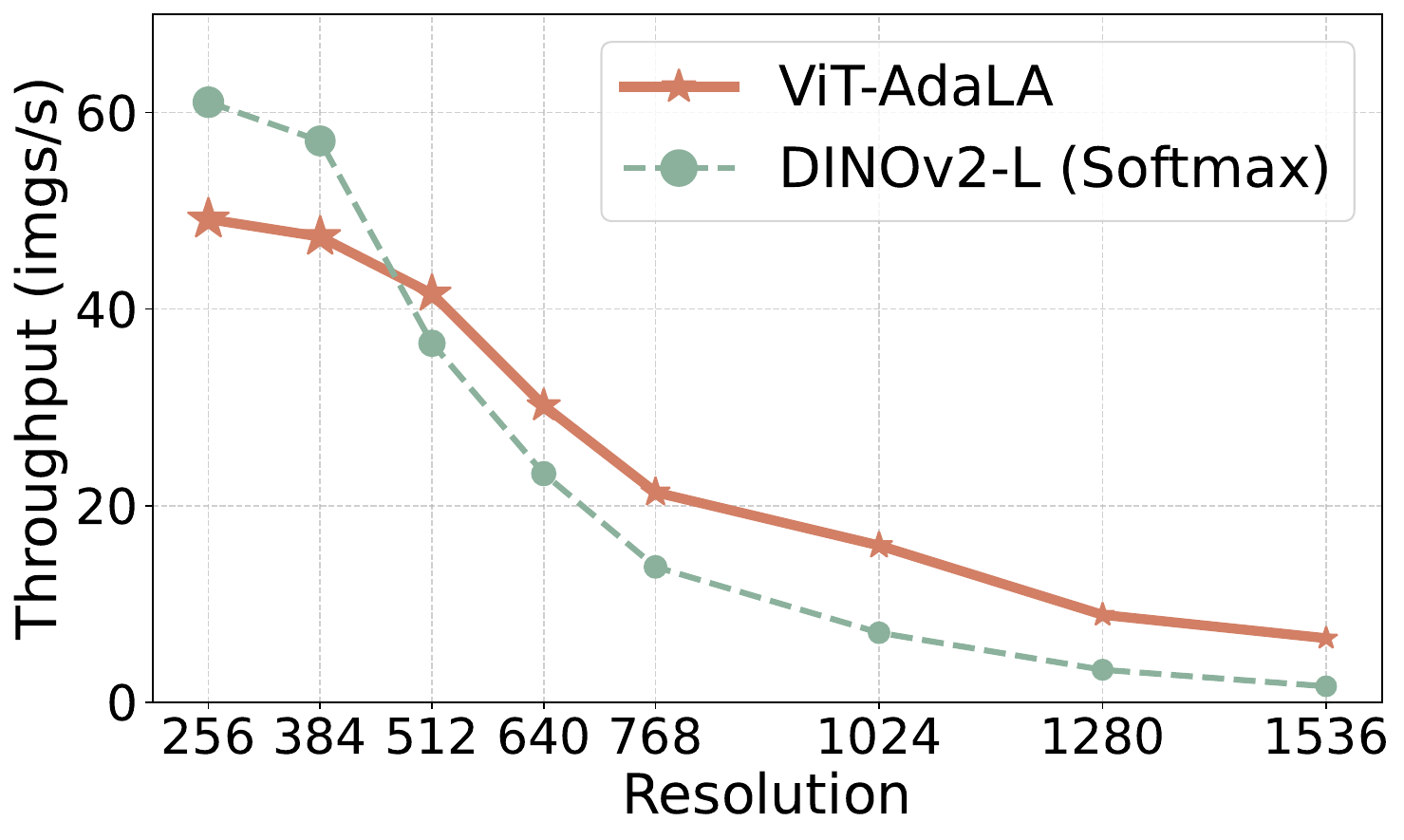}
\centerline{\small(b) Throughput}
\end{minipage}
\hfill
\begin{minipage}[b]{0.32\linewidth}
\centering
\includegraphics[width=\linewidth]{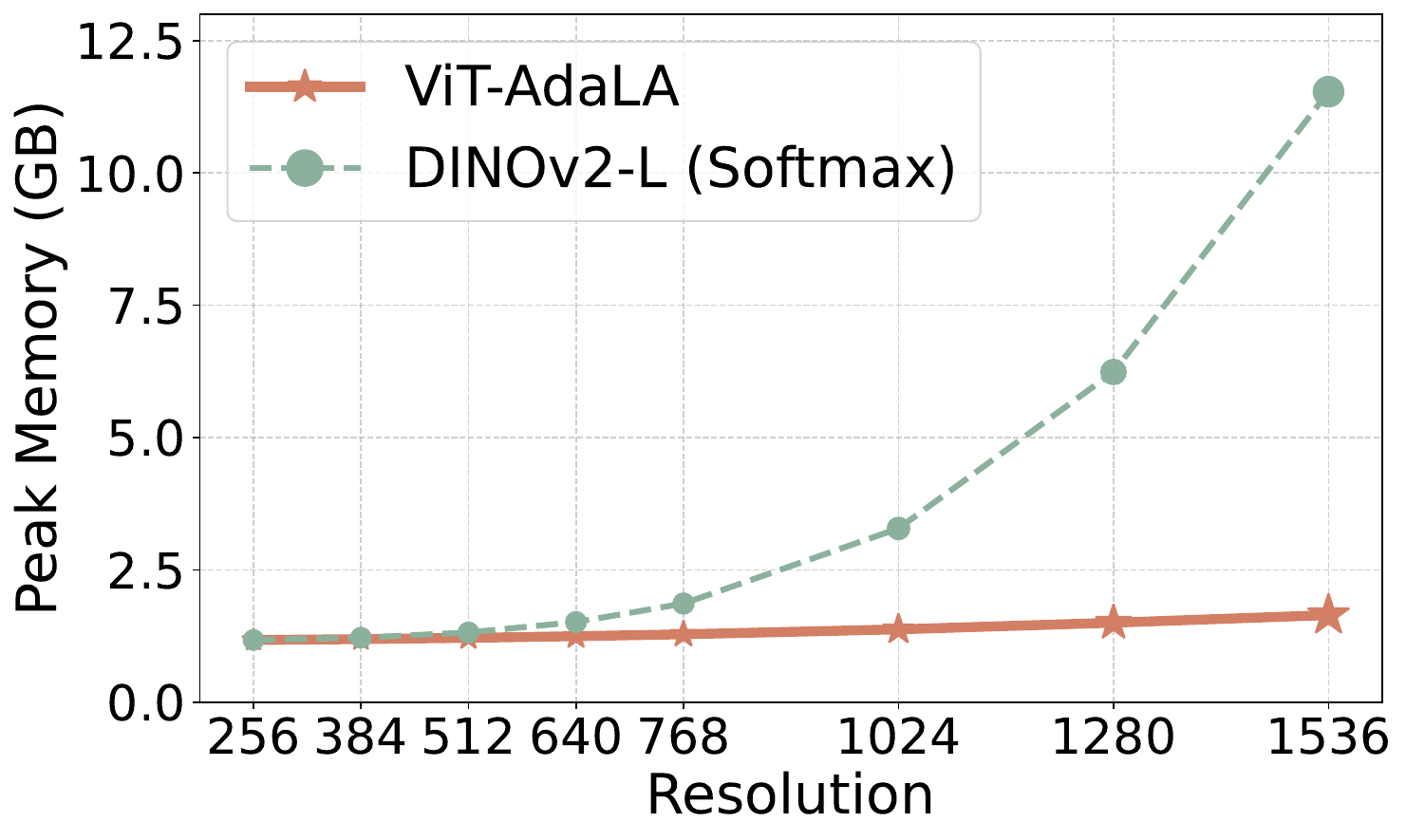}
\centerline{\small(c) Peak memory, batch size one}
\end{minipage}
\caption{Resolution scalability on Cityscapes: mIoU, throughput, and peak memory.}
\label{fig:resolution_scalability_efficiency_comparison}
\end{figure*}

\begin{table}[!htbp]
\centering
\caption{Stage 1/2 ablation on ADE20K (DINOv2-L) and SAM-HQ (SAM-H).}
\label{tab:stage_ablation}
\small
\setlength{\tabcolsep}{14pt}
\begin{tabular}{cccc}
\toprule
Stage 1 & Stage 2 & ADE20K mIoU & SAM-HQ IoU \\
\midrule
$\times$ & $\times$ & 22.92 & 63.65 \\
\checkmark & $\times$ & 19.37 & 72.00 \\
$\times$ & \checkmark & 52.46 & 73.47 \\
\checkmark & \checkmark & \textbf{55.55} & \textbf{77.48} \\
\midrule
\multicolumn{2}{c}{Softmax teacher} & 56.73 & -- \\
\bottomrule
\end{tabular}
\end{table}

\textbf{Stage 1: Attention Alignment.} Stage 1 improves the Stage-2 result by 3.09 mIoU on ADE20K and 4.01 IoU on SAM-HQ. Figure~\ref{fig:loss_comparison_w_wo_stage1} shows a faster decrease in Stage-2 feature-MSE with Stage 1 initialization.

Stage 1's standalone effect is task-dependent: on ADE20K it obtains 19.37 mIoU, below the unaligned baseline (22.92), whereas on SAM-HQ it raises IoU from 63.65 to 72.00 (+8.35). These results support Stage 1 as a complementary initialization that can also improve foreground segmentation independently.

Figure~\ref{fig:all_attn_losses_stage1} supports vanilla linear attention as our default through its lower Stage-1 alignment loss.

\textbf{Stage 2: Feature Alignment.} \emph{Recovering pretrained compatibility.} With the attention architecture fixed, Stage 2 alone raises ADE20K mIoU from 22.92 to 52.46 (recovering 87.4\% of the teacher gap) and SAM-HQ IoU from 63.65 to 73.47 (Table~\ref{tab:stage_ablation}). Direct-replacement failure therefore need not imply inadequate linear-attention capacity; it can reflect disrupted compatibility with pretrained blocks.

\emph{Local fidelity is not sufficient.} On ADE20K, Stage 1 alone reduces mIoU by 3.55, yet adds 3.09 with Stage 2: its transfer benefit is conditional on global adaptation. Stage 1 fits attention on teacher inputs; the student receives features altered by preceding replacements. Stage 2 aligns final features under the student's actual forward pass, letting the whole encoder co-adapt without layer-by-layer equality (Appendix~\ref{sec:spatia_error}). It converts local initialization into useful end-to-end transfer.

\emph{Recovery is transferable.} After Stages 1--2, frozen-backbone transfer retains 98.4\% of teacher ImageNet-1K accuracy and 99.7\% on Flowers-102 (Table~\ref{tab:frozen_transfer}); held-out Flowers-102 CKA rises from 0.196 to 0.993 (Table~\ref{tab:heldout_geometry}). Recovery is thus visible on unseen data without downstream backbone updates. With the same full pipeline, Nystr\"omformer and Monarch gain 13.44 and 9.95 mIoU (Table~\ref{tab:matched_attention_alignment}), supporting a shared representation-transfer objective across attention modules. Figure~\ref{fig:ade20k_performance_across_stage2_epochs} shows Stage-2 progress.

\begin{tcolorbox}[
    enhanced,
    colback=gray!5,          
    colframe=teal!30!black,  
    left=3mm,right=3mm,
    top=1mm,bottom=1mm,
    boxrule=0.8pt,           
    arc=4pt,                 
    drop shadow              
]
\itshape
VFM linearization should preserve the encoder's transferable representation, not merely approximate its individual attention operators.
\end{tcolorbox}
\subsubsection{Resolution Scalability Analysis}\label{sec:resolution_scalability}
Although adapted at $512^2$, ViT-AdaLA transfers to Cityscapes resolutions from $256^2$ to $1536^2$ (Figure~\ref{fig:resolution_scalability_efficiency_comparison}). It maintains strong mIoU while improving throughput and memory efficiency at high resolution. At $256^2$ and batch size one, however, softmax is faster; the linearized model overtakes it by $512^2$. Matched throughput measurements are given in Appendix~\ref{sec:efficiency_regime}.

\subsubsection{Directly Adapting the Task Model}\label{sec:task_teacher_comparison}
Figure~\ref{fig:directly_adapting_task_model} compares adapting a task-trained teacher with adapting the general-purpose VFM. The task-trained teacher reaches 56.04 mIoU on ADE20K, compared with 55.55 for the VFM teacher, suggesting that task-specific representations can also be transferred. Both settings use full downstream fine-tuning after conversion; this is not a frozen-backbone evaluation.

\begin{table}[!t]
\centering
\caption{DINOv2-L frozen-backbone transfer after Stage 2. Matched task heads are trained; retained quality is the student-to-teacher ratio. ADE20K values are approximate.}
\label{tab:frozen_transfer}
\small
\setlength{\tabcolsep}{5pt}
\begin{tabular}{lcccc}
\toprule
Task & Head / frozen (M) & Softmax & ViT-AdaLA & Retained \\
\midrule
ImageNet-1K linear probe & 1.0M / 304M & 82.66 & 81.30 & 98.4\% \\
Flowers-102 linear probe & 0.10M / 304M & 99.71 & 99.37 & 99.7\% \\
ADE20K Mask2Former & 14.8M / 304M & $\sim$50.0 & $\sim$49.5 & $\sim$99.0\% \\
\bottomrule
\end{tabular}
\end{table}


\subsubsection{Frozen-Backbone Transfer}\label{sec:frozen_backbone_transfer}
We freeze all 304M DINOv2-L backbone weights after Stage 2 and train matched task heads (Table~\ref{tab:frozen_transfer}). ViT-AdaLA retains 98.4\% of the softmax teacher's ImageNet-1K accuracy, 99.7\% on held-out Flowers-102, and approximately 99.0\% ADE20K mIoU. Thus, Stage 3 full-backbone fine-tuning is not required for these tasks; the transferred features remain useful with only head training.

\begin{wraptable}{r}{0.54\textwidth}
\vspace{-25pt}
\centering
\captionsetup{font=small}
\caption{Full-network versus QKV-only tuning on CLIP-L and DINOv2-L at $512^2$ resolution.}
\label{tab:clip_dino_tuneqkv}
\small
\setlength{\tabcolsep}{3pt}
\begin{tabular}{llcc}
\toprule
Backbone & Tuning & \shortstack{ImageNet-1K\\Acc.} & \shortstack{ADE20K\\mIoU} \\
\midrule
CLIP-L & Full network & 85.5 & 42.85 \\
CLIP-L & QKV only & 82.5 & 35.41 \\
\midrule
DINOv2-L & Full network & 86.0 & 55.55 \\
DINOv2-L & QKV only & 76.4 & 48.82 \\
\bottomrule
\end{tabular}
\vspace{-8pt}
\end{wraptable}

\subsubsection{Effect of QKV-Only Tuning}\label{sec:qkv_tuning_ablation}
Table~\ref{tab:clip_dino_tuneqkv} compares QKV-only tuning with full-network tuning. The latter improves ImageNet-1K accuracy by 3.0 points on CLIP-L and 9.6 points on DINOv2-L, and ADE20K mIoU by 7.44 and 6.73 points, respectively. The gap indicates that QKV-only updates, while useful for local attention alignment in Stage 1, do not suffice for the complete adaptation; updating the wider network recovers substantially more task performance.

\vspace{-5pt}
\section{Conclusion}\vspace{-5pt}
We introduce ViT-AdaLA, a framework that adapts Vision Foundation Model (VFM) priors into linearized ViTs without large-scale pretraining. Our three-stage alignment process effectively distills priors, accelerates convergence, and ensures robust resolution scalability. This paradigm offers a fresh perspective for linearization research, paving the way for high-resolution image applications.
Future research can explore more efficient and effective linearization architectures.
\begin{ack}
This project is partially supported by the Adobe gift funding program and NSF grant \#2429836 under the NSF Responsible Design, Development, and Deployment of Technologies (ReDDDoT) program. The ReDDDoT program is jointly sponsored by NSF and the Ford Foundation, the Patrick J. McGovern Foundation, Pivotal Ventures, the Eric and Wendy Schmidt Fund for Strategic Innovation, and Siegel Family Endowment.
\end{ack}

\bibliographystyle{plainnat}
\bibliography{references}

\clearpage
\appendix

\startcontents[appendix]
\section*{Appendix Contents}
{\hypersetup{hidelinks}\small\printcontents[appendix]{}{1}{\setcounter{tocdepth}{2}}}
\clearpage

\section{Experiments}

\subsection{Experimental Setup and Baselines}
We summarize the training configuration and the reproduced baseline implementations.

\subsubsection{Configurations of ViT-AdaLA under Different VFMs}\label{sec:hyper-parameter_configs}
We train Stage 1 on COCO for four epochs (batch size 32 per GPU, fixed learning rate $10^{-2}$) and Stage 2 on ImageNet-22K for 10--30 epochs depending on the backbone (batch size 16 per GPU, initial learning rate $10^{-3}$ with polynomial decay and a 0.1 backbone learning-rate multiplier). Both stages use AdamW, eight H100 GPUs, and $512\times512$ inputs with random cropping and color jitter. Table~\ref{tab:training_stage_configuration} gives the per-backbone settings.

\begin{table}[!htbp]
\caption{Hyperparameters of different training stages and VFMs.}\label{tab:training_stage_configuration}
\centering
\renewcommand{\arraystretch}{1.5} 
\resizebox{0.85\textwidth}{!}{%
\begin{tabular}{l c cccc cc}
\toprule
 & \textbf{Stage 1} & \multicolumn{4}{c}{\textbf{Stage 2}} & \multicolumn{2}{c}{\textbf{Stage 3}} \\
\cmidrule(lr){2-2} \cmidrule(lr){3-6} \cmidrule(lr){7-8}
\multirow{2.5}{*}{Epochs} & \multirow{2.5}{*}{4} & DINOv2 & SigLIP & CLIP & IN-1K & Cls & seg \\
\cmidrule(lr){3-6} \cmidrule(lr){7-8}
 & & 30 & 10 & 15 & 20 & 10 & 20 \\
\midrule
Batch size & 32 & \multicolumn{4}{c}{16} & \multicolumn{2}{c}{8} \\
\midrule
Learning rate & 1e-2 & \multicolumn{4}{c}{1e-3} & \multicolumn{2}{c}{1e-4} \\
\midrule
Weight decay & \multicolumn{7}{c}{0.05} \\
\midrule
Learning schedule & fixed & \multicolumn{6}{c}{Polynomial (decay factor of 0.9)} \\
\midrule
Optimizer & \multicolumn{7}{c}{AdamW} \\
\midrule
Datasets & COCO & \multicolumn{4}{c}{ImageNet22K} & ImageNet1K & ADE20K/Cityscapes \\
\midrule
Heads & N/A & \multicolumn{4}{c}{N/A} & Linear & Mask2former \\
\midrule
$\lambda$ & N/A & 5 & 1 & 4000 & 1 & \multicolumn{2}{c}{N/A} \\
\bottomrule
\end{tabular}%
}
\end{table}

\subsubsection{Details of Compared Baselines}
In our experiments, we compare with multiple linear attention baselines including Hedgehog~\cite{zhanghedgehog}, LoLCATS \cite{zhanglolcats}, Linformer \cite{wang2020linformer}, Performer \cite{choromanskirethinking}, Cosformer \cite{qincosformer}, Nystromformer \cite{xiong2021nystromformer} and Monarch Attention \cite{yaras2025monarchattention}. We reproduce all of these methods according to their perspective publicly available codebase. Below are the details about these baselines:

\textbf{Hedgehog}: Hedgehog \cite{zhanghedgehog} proposes two trainable linear attention feature maps $\phi(\cdot)$ after queries and keys. It then learns these two feature maps to mimic standard attention. By employing a distillation loss to train these maps directly against the original attention scores, Hedgehog bridges the performance gap between models. 
This approach allows the model to maintain linear complexity while recovering most power of the original model.

\textbf{LoLCATS}: LoLCATS \cite{zhanglolcats} adopts a two-stage methodology to linearize LLMs, converting their quadratic complexity attention into a faster, subquadratic form without the massive cost of retraining from scratch. In the first stage, attention transfer, the model utilizes learnable feature maps (inspired by the Hedgehog architecture) trained via the MSE loss to approximate the final attention outputs, rather than the attention matrix itself. The second stage, low-rank adjusting, employs LoRA to tune the $Q$, $K$, $V$, and $O$ projections followed by task-specific supervised fine-tuning. This methodology scales exceptionally well, delivering high performance and efficiency across  decoder-based LLMs.

\textbf{Monarch Attention}: Monarch Attention \cite{yaras2025monarchattention} introduces a zero-shot method to convert standard softmax attention into a hardware-efficient, subquadratic form using Monarch matrices. Different from previous methods that require extensive retraining, Monarch Attention allows the plug-and-play replacement of attention layers in pretrained Transformers, maintaining high accuracy. This is achieved by the mathematical property that any dense matrix can be approximated by Monarch matrices \cite{dao2022monarch}. By decomposing the attention operation into these structured components, they achieve a complexity of $\mathcal{O}(n\sqrt{n})$. The proposed method proves effective across diverse architectures, including ViTs, various language model configurations (encoder-based and encoder-decoder), and Diffusion Transformers.

\textbf{Nystr\"{o}mformer}: Nystr\"{o}mformer \cite{xiong2021nystromformer} proposes to approximate the self-attention using the Nystr\"{o}m method \cite{wang2013improving} by sampling a subset of columns and rows. The core of this method involves a small number of ``landmark'' points to reconstruct the full, inherently low-rank attention matrix. This method further improves accuracy with an iterative Moore-Penrose pseudoinverse approximation and residual connections that stabilize training. Nystr\"{o}mformer allows the model to process sequences with thousands of tokens efficiently while maintaining performance comparable performance comparable to standard BERT~\cite{devlin2019bert} models. In our experiments, we set the number of ``landmark'' points to 128, which is the head dimension.

\textbf{Linformer}: Linformer \cite{wang2020linformer} introduces a low-rank approximation of self-attention based on the empirical observation that attention matrices have low intrinsic rank. It achieves linear complexity by projecting the key and value sequences into a lower-dimensional space using learned projection matrices, reducing the size of the attention computation from $N\times N$ to $N\times k$ with $k \ll N$. These projections are shared across heads and layers, allowing efficient scaling to long sequences with minimal overhead. Experiments show that Linformer matches or closely approaches standard Transformer performance on NLP tasks while dramatically lowering memory and computation costs. In our experiments, we set $k=32$.

\textbf{Performer}: Performer \cite{choromanskirethinking} rethinks self-attention by replacing the softmax attention kernel with a random feature approximation (FAVOR+), enabling exact linear time attention in expectation. By mapping queries and keys into a positive random feature space, the attention computation is reformulated as a sequence of associative matrix products, avoiding explicit $N\times N$ attention matrices. This design preserve the probabilistic interpretation of softmax attention while strong numerical stability and unbiased estimates. Experiments show that Performer scales effectively to very long sequences with competitive accuracy, though performance depends on the number of random features and can degrade when approximation variance is high.

\textbf{Cosformer}: Cosformer \cite{qincosformer} revists the softmax operation in self-attention and proposes replacing it with a cosine-based reweighting mechanism that enables linear-time attention. The authors argue that softmax's effectiveness comes from two key properties: nonnegativity of the attention matrix and a non-linear reweighting that concentates attention distributions. Specifically, Cosformer utilizes the ReLU-based kernel to ensure non-negativity and cosine-based reweighting to introduce a ``locality bias'' that stabilizes training and focuses the model on more relevant local correlations. In experiments, Cosformer evaluates its effectiveness  on encoder-based and decoder-based language models.




\subsection{Efficiency and Alignment Diagnostics}
We examine resolution scaling and the local-to-global alignment errors that motivate Stage 2.

\subsubsection{Additional Resolution-Dependent Inference Measurements}\label{sec:efficiency_regime}
Table~\ref{tab:resolution_crossover} supplements the resolution curves in Figure~\ref{fig:resolution_scalability_efficiency_comparison} with end-to-end throughput on one H100. At $256^2$ and batch size one, the linearized model is slower; it overtakes softmax by $512^2$ and gains substantially at $1024^2$. Wider models may require longer sequences, but a width-dependent crossover was not measured.
\begin{table}[!htbp]
\centering
\caption{End-to-end throughput (images/s) for DINOv2-L on one H100. The $256^2$ batch-size results are also shown in Table~\ref{tab:dinov2l_efficiency_batch_size}.}
\label{tab:resolution_crossover}
\small
\begin{tabular}{cccc}
\toprule
Resolution & Batch size & Softmax & ViT-AdaLA \\
\midrule
$256^2$ & 1 & 62.83 & 49.28 \\
$256^2$ & 16 & 252.29 & 260.54 \\
$512^2$ & 1 & 36.52 & 41.56 \\
$1024^2$ & 1 & 7.07 & 15.95 \\
\bottomrule
\end{tabular}
\end{table}

\FloatBarrier

\subsubsection{Layer Sensitivity}\label{sec:layer_sensitivity}
\begin{figure}
    \centering
    \includegraphics[width=0.8\linewidth]{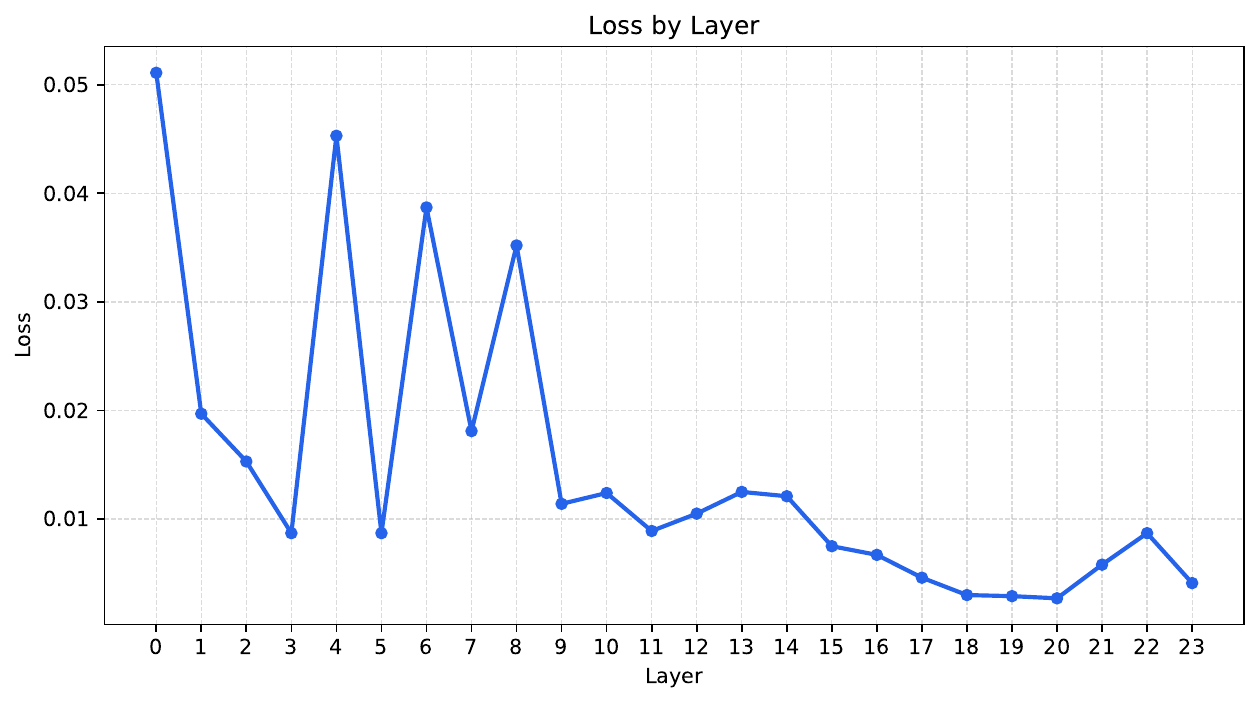}
    \caption{Layer-wise loss of Stage1 for DINOv2-L.}
    \label{fig:layer-wise-loss-stage1}
\end{figure}

To investigate layer-wise sensitivity, we measured the approximation loss for each layer of DINOv2-L (Stage 1) in Fig. \ref{fig:layer-wise-loss-stage1}.

The isolated per-layer losses range from 0.0027 to 0.0511, with a sum of 0.3546. Replacing all layers gives a final-output discrepancy of 2.2839. Because these quantities are measured at different outputs, the ratio is a diagnostic of compounded representation drift rather than a layer-independent amplification coefficient.

The diagnostic motivates Stage 2: matching each attention block locally does not by itself preserve the representation of the full pretrained vision encoder.

\subsubsection{Accumulated Representation Drift in ViTs}\label{sec:spatia_error}

Tab. \ref{tab:dinov2l_reconstruction_loss_stage1} compares the L2 reconstruction loss of individual layers after Stage 1, \textit{i.e.}, Layer-wise Distillation, against the final output loss when all layers are replaced with linear attention simultaneously.

\begin{table}[t]
\centering
\caption{Reconstruction loss for DINOv2-L after Stage 1.}
\label{tab:dinov2l_reconstruction_loss_stage1}
\resizebox{\linewidth}{!}{
\begin{tabular}{lccc}
\toprule
\textbf{Metric} 
& \textbf{Mean Layer-wise Loss (0--23)} 
& \textbf{Sum of Layer-wise Losses} 
& \textbf{Final Output Loss (Accumulated)} \\
\midrule
Value & 0.0148 & 0.3546 & 2.2839 \\
\bottomrule
\end{tabular}
}
\end{table}

\noindent\textit{Note: Individual layer losses range from 0.0027 to 0.0511.}

\begin{itemize}
    \item \textbf{The compounding effect.} 
    The sum of isolated layer losses is 0.3546, while the final-output discrepancy after replacing all layers is 2.2839. These losses have different output locations and scales, so their ratio should not be interpreted as a formal error-propagation coefficient. The result nevertheless shows that local alignment alone does not control final representation drift.
\end{itemize}

\paragraph{Local mismatch versus propagated drift.}
Let $f_i$ and $g_i$ denote the complete teacher and student blocks, including attention, normalization, feed-forward, and residual operations; the last mapping also includes the final encoder normalization. For a fixed image, their input token features are $\mathbf{h}_i$ and $\hat{\mathbf{h}}_i$, respectively, flattened for the following analysis. Define input drift $\boldsymbol{\delta}_i=\hat{\mathbf{h}}_i-\mathbf{h}_i$ and the block mismatch evaluated on the teacher input, $\boldsymbol{\epsilon}_i=g_i(\mathbf{h}_i)-f_i(\mathbf{h}_i)$. An exact decomposition is
\begin{equation}\label{eq:alignment_error_decomposition}
\boldsymbol{\delta}_{i+1}
=\underbrace{g_i(\mathbf{h}_i+\boldsymbol{\delta}_i)-g_i(\mathbf{h}_i)}_{\text{response to shifted input}}
+\underbrace{\boldsymbol{\epsilon}_i}_{\text{teacher-input block mismatch}}.
\end{equation}
Stage 1 fits attention outputs on shared teacher inputs. It therefore reduces a source of block mismatch, but neither its attention-output loss nor an isolated block loss directly measures the first term in Eq.~\ref{eq:alignment_error_decomposition}.

For sufficiently small drift, a first-order expansion with $\mathbf{J}_i=\left.\partial g_i/\partial\mathbf{h}\right|_{\mathbf{h}_i}$ gives
\begin{equation}\label{eq:alignment_error_transport}
\boldsymbol{\delta}_{i+1}\approx\mathbf{J}_i\boldsymbol{\delta}_i+\boldsymbol{\epsilon}_i,
\qquad
\boldsymbol{\delta}_{K}\approx\mathbf{P}_{K,0}\boldsymbol{\delta}_0
+\sum_{i=0}^{K-1}\mathbf{P}_{K,i+1}\boldsymbol{\epsilon}_i,
\end{equation}
where $\mathbf{P}_{K,j}=\mathbf{J}_{K-1}\cdots\mathbf{J}_j$ for $j<K$ and $\mathbf{P}_{K,K}=\mathbf{I}$. The initial drift $\boldsymbol{\delta}_0$ is zero when the input embeddings are shared, but can change when Stage 2 updates the student's embedding parameters. Each local mismatch is transformed by the remaining blocks before contributing to final drift. Consequently, minimizing individual local errors is not equivalent to minimizing their combined effect at the encoder output: downstream sensitivities and interactions between error directions also matter.

\paragraph{Role of Stage 2.}
Stage 2 directly supervises the final representation under the student's actual intermediate inputs. Updating the whole encoder allows both block mismatches and their propagation to change jointly, rather than restricting adaptation to QKV weights or requiring layer-by-layer equality. Its objective preserves the teacher's external feature mapping while leaving the student's internal computation free to adapt. The recovery from 22.92 to 52.46 mIoU with Stage 2 alone (Table~\ref{tab:stage_ablation}) is consistent with this distinction; it does not establish which Jacobians changed or quantify amplification. Eq.~\ref{eq:alignment_error_transport} is a local explanatory approximation, not an empirical measurement, and can be inaccurate for large drift.

The same shifted-input phenomenon occurs in deep Transformers generally. In a bidirectional ViT it can disturb relationships among spatial patches and the final feature geometry; it is not exclusive to vision models.

The Stage-2 output is also evaluated directly by frozen-head transfer (Table~\ref{tab:frozen_transfer}) and held-out representation metrics (Table~\ref{tab:heldout_geometry}). We do not infer from this vision-only experiment that LLMs lack accumulated residual errors.

\subsubsection{Activation Choices for Linear Attention}\label{sec:activation_choices}
\begin{figure}
    \centering
    \includegraphics[width=0.53\linewidth]{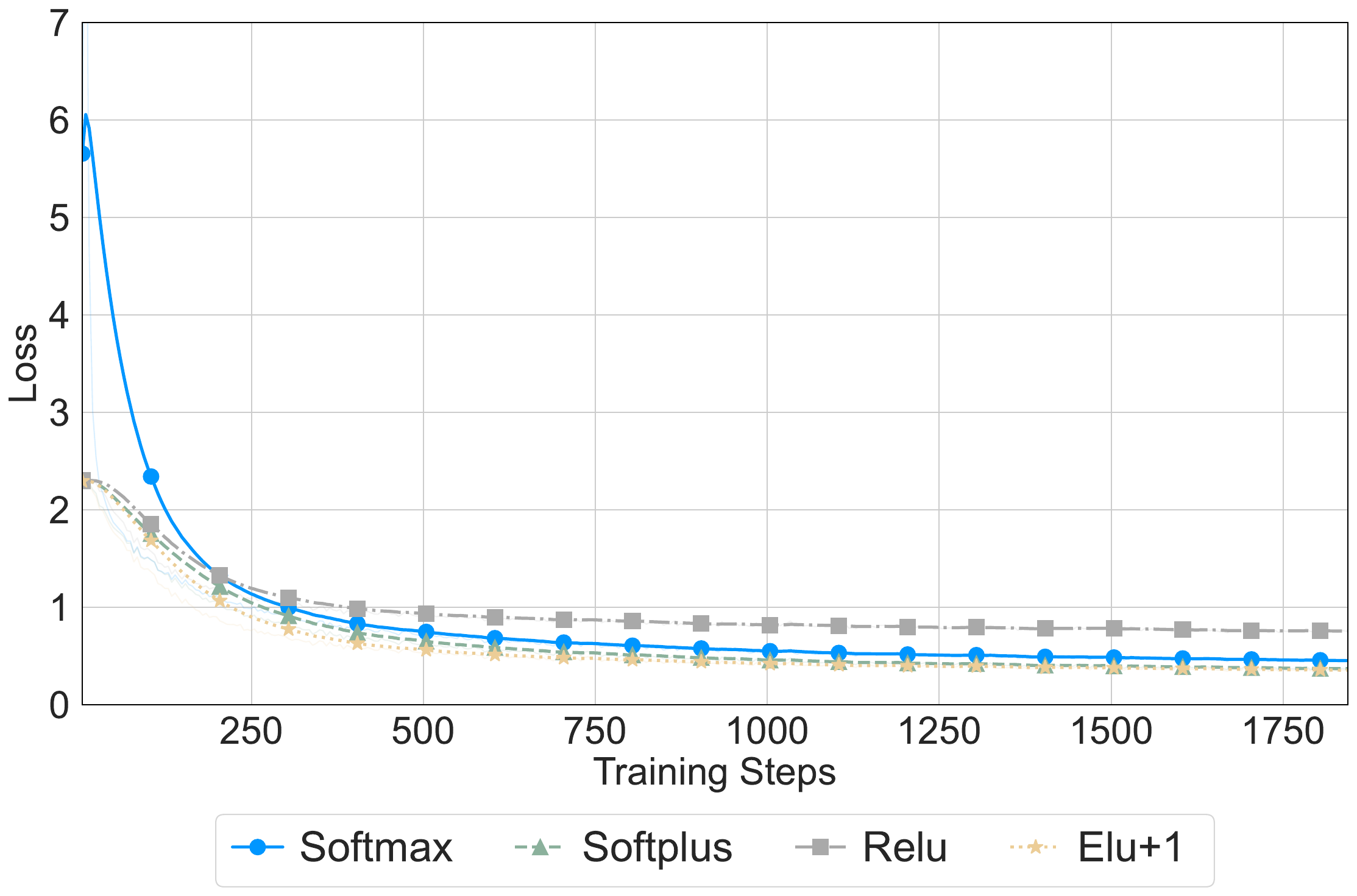}
    \caption{Stage 1 loss comparison across linear-attention activation functions. We compare four activations: softmax, softplus, ReLU, and ELU+1.}
    \label{fig:activation_linear_attention}
\end{figure}

Fig. \ref{fig:activation_linear_attention} shows the influence of activation functions to the Stage 1 loss. Kernel-based linear attention approximation applys a non-negative feature map $\phi(\cdot)$ to the query and key matrices. We compare four different variants, \textit{i.e.}, softmax, softplus, relu and ELU+1.

While linear attention aims to approximate this, the standard self-attention uses the softmax activation to normalize scores:
$$
\mathrm{Softmax}(x)_i=\frac{e^{x_i}}{\sum_je^{x_j}}.
$$
In kernelized linear attention, an exponential feature map $\phi(x)=\mathrm{exp}(x)$ is sometimes used to mimic this behavior, though it can be numerically unstable without proper scaling.

ReLU is a popular choice for linear attention due to its computational simplicity and ability to induce sparsity:
$$
\phi(x)=\mathrm{ReLU}(x)=\max(0, x).
$$
By mapping negative values to zero, it ensures the non-negativity required for the associative property of matrix multiplication in linear attention.

Softplus~\cite{pmlr-v15-glorot11a} is a smooth, differentiable approximation of ReLU:
$$
\phi(x)=\mathrm{Softplus}(x)=ln(1+e^x).
$$
This activation is strictly positive and provides continuous gradients throughout the entire domain, which can lead to more stable training trajectories compared to ReLU.
Commonly used in the original linear attention \cite{katharopoulos2020transformers}, the $\mathrm{ELU}(x)+1$ feature map ensures that the output is always positive:
$$
\phi(x)=\mathrm{ELU}(x)+1=\begin{cases} 
x + 1 & \text{if } x > 0 \\ 
e^x & \text{if } x \le 0 
\end{cases}.
$$
This function combines the linear behavior for positive inputs with a smooth exponential decay toward zero for negative inputs.

As illustrated in Fig. \ref{fig:activation_linear_attention}, $\mathrm{ELU+1}$ feature map consistently achieves a lower training loss compared to other activation functions. Given its superior convergence characteristics and numerical stability, we select $\mathrm{ELU+1}$ as the default activation for our final architecture. 
\FloatBarrier

\subsection{Training Cost and Model Ablations}
We report adaptation time and compare model sizes and scratch training.

\subsubsection{Training Time Comparison}
Table~\ref{tab:training_time} reports measured per-epoch wall times on eight H100 GPUs. Stage 1 is brief, while Stage 2 dominates the one-time adaptation cost; aggregate GPU-hours are given in Table~\ref{tab:stage_compute}.
\begin{table}[htbp]
  \centering
  \caption{Comparison of training efficiency across various Visual Foundation Models (VFMs) during Stages 1 and 2. Stage 1 is conducted using the COCO dataset, while Stage 2 utilizes ImageNet-22K for large-scale pretraining. All the experiments are conducted on 8 H100s.}
  \label{tab:training_time}
  \begin{tabular}{lcc}
    \toprule
    Stages & Stage 1 Time (per epoch) & Stage 2 Time (per epoch) \\
    \midrule
    DINOv2-L & 4min 37s & 10h 48min 46s \\
    SigLIP-L & 4min 19s & 11h 22min 15s \\
    CLIP-L   & 5min 56s & 11h 09min 21s \\
    IN1K-L   & 4min 34s & 10h 02min 54s \\
    \midrule
    \textbf{Mean} & \textbf{4min 51s} & \textbf{10h 55min 41s} \\
    \bottomrule
  \end{tabular}
\end{table}

\subsubsection{Performance on Smaller Size VFMs}
Apart from the results on DINOv2-L, we also report experiments on DINOv2-B in Tab.~\ref{tab:dinov2-base-experiments} for comparison. The results show that the performance gap between the original softmax model and our ViT-AdaLA is small, demonstrating the effectiveness of our method across different model sizes.
\begin{table}[!htbp]
\centering
\caption{Performance comparison using DINOv2-B on ADE20K, Cityscapes and ImageNet-1K. We pretrained our ViT-AdaLA (DINOv2-B) for 40 epochs.}
\label{tab:dinov2-base-experiments}
\begin{tabular}{lccc}
\toprule
Method & ADE20K (mIoU) & Cityscapes (mIoU) & IN1K (Acc.) \\
\midrule
Softmax (DINOv2-B) & 53.93 & 79.73 & 84.93 \\
ViT-AdaLA (DINOv2-B) & 51.46 & 74.60 & 84.37 \\
\bottomrule
\end{tabular}
\end{table}

\subsubsection{Comparison with Training from Scratch}\label{sec:scratch_comparison}
\begin{table}[!htbp]
\centering
\caption{Reference comparison for IN-1K ViT-L. Image counts cover ImageNet-22K Stage 2 for ViT-AdaLA and ImageNet-1K pretraining for the scratch baseline. Data, teacher supervision, and compute are not matched.}
\label{tab:train-from-scratch-comparison}
\small
\begin{tabular}{lcccc}
\toprule
Method & Images seen & ADE20K & Cityscapes & IN-1K \\
\midrule
Softmax (IN-1K-L) & -- & 50.83 & 74.86 & 85.66 \\
ViT-AdaLA (IN-1K-L) & 283.9M & 49.75 & 72.40 & 84.23 \\
Linear ViT from scratch & 256.2M & 25.80 & 45.09 & 65.72 \\
\bottomrule
\end{tabular}
\end{table}

We train a vanilla linear-attention ViT on ImageNet-1K for 200 epochs and compare its downstream results with an IN-1K-pretrained ViT-L adapted on ImageNet-22K for 20 Stage-2 epochs (Table~\ref{tab:train-from-scratch-comparison}). ImageNet-22K has approximately 14.20M images per epoch versus 1.28M for ImageNet-1K: these runs see roughly 283.9M and 256.2M training images, respectively. They are of comparable order in image exposures but are neither data- nor compute-matched: our method also uses a pretrained softmax teacher. The scratch curve had plateaued by 200 epochs and additional training gave no meaningful improvement. We use this run as a reference for the benefit of reusing the pretrained VFM, not as proof that adaptation always costs less.

\FloatBarrier
\subsection{Task Results and Batch-Size Scaling}
We report additional classification and segmentation results, followed by batch-size efficiency.

\subsubsection{Experiments on Classification Tasks}
Tab. \ref{tab:imagenet_benchmarks_other_VFMs} presents the classification benchmarks for various linear attention models based on CLIP-L. From the results, it can be observed that approximation-based linear attention methods, \textit{i.e.}, Monarch and Nystr\"omformer, significantly outperform other baselines. This demonstrates that a good approximation of softmax is advantageous. However, these two methods still cannot match the performance of our ViT-AdaLA or even the Stage 2 variant. This performance gap suggests that linearization-based approaches are more effective at distilling prior knowledge from pre-trained VFMs compared to traditional training-from-scratch paradigms.

\begin{table*}[!htbp]
\centering
\caption{Top-1 fine-tuning accuracy comparison on ImageNet-1K under CLIP-L with multiple linear attention baselines. Native/direct conversions use the same downstream full fine-tuning budget; see Table~\ref{tab:baseline_protocol}. The classification head is a single linear layer for all the methods. }
\label{tab:imagenet_benchmarks_other_VFMs}
\setlength{\tabcolsep}{4.5pt}
\renewcommand{\arraystretch}{1.3}
\scalebox{0.85}{
\begin{tabular}{p{3cm} c c c c c c c}
\toprule
& \multirow{2}{*}{\textbf{Backbone}} & \multirow{2}{*}{\textbf{Res.}} & \textbf{Params} & {\textbf{FLOPS}} & \textbf{Peak Mem.} & {\textbf{Throughput}} & \textbf{Top-1 Acc.} \\
& & & \textbf{(M)} &\textbf{(G)} & \textbf{(GB)} & \textbf{(imgs/s)} & \textbf{(\%)} \\
\midrule
\rowcolor{gray!10} Softmax \cite{oquab2024dinov2} & CLIP-L & $512^2$ & 304.20 & 310.60 & 1.3181 & 36.52 & 86.8 \\
Hedgehog \cite{zhanghedgehog}  & CLIP-L & $512^2$ & 305.77 & 265.41\textcolor{darkgreen}{$_{\downarrow 14.5\%}$} & 1.2221\textcolor{darkgreen}{$_{\downarrow 7.3\%}$} & 37.44\textcolor{darkgreen}{$_{\uparrow 2.5\%}$}  & 67.4 \\
LoLCATS \cite{zhanglolcats}  & CLIP-L & $512^2$ & 305.77 & 265.41\textcolor{darkgreen}{$_{\downarrow 14.5\%}$} & 1.2221\textcolor{darkgreen}{$_{\downarrow 7.3\%}$} & 37.44\textcolor{darkgreen}{$_{\uparrow 2.5\%}$} & 67.21 \\
Linformer \cite{wang2020linformer} & CLIP-L & $512^2$ & 305.77 & 262.19\textcolor{darkgreen}{$_{\downarrow 15.6\%}$} & 1.2221\textcolor{darkgreen}{$_{\downarrow 7.3\%}$} & 45.41\textcolor{darkgreen}{$_{\uparrow 25.3\%}$} & 56.8 \\
Performer \cite{choromanskirethinking} & CLIP-L & $512^2$ & 304.20 & 265.41\textcolor{darkgreen}{$_{\downarrow 14.6\%}$} & 1.2164\textcolor{darkgreen}{$_{\downarrow 7.7\%}$} & 36.48\textcolor{darkgreen}{$_{\uparrow 0.1\%}$} & 69.9 \\
Cosformer \cite{qincosformer} & CLIP-L & $512^2$ & 304.20 & 265.41\textcolor{darkgreen}{$_{\downarrow 14.6\%}$} & 1.2226\textcolor{darkgreen}{$_{\downarrow 7.2\%}$} & 39.55\textcolor{darkgreen}{$_{\uparrow 8.2\%}$} & 70.6 \\
Nystr\"{o}mformer  \cite{xiong2021nystromformer} & CLIP-L & $512^2$ & 304.20 & 265.20\textcolor{darkgreen}{$_{\downarrow 14.6\%}$} & 1.2163\textcolor{darkgreen}{$_{\downarrow 7.7\%}$} & 18.29\textcolor{darkred}{$_{\downarrow 50.8\%}$} & 80.7 \\
Monarch \cite{yaras2025monarchattention} & CLIP-L & $512^2$ & 304.20 & 269.44\textcolor{darkgreen}{$_{\downarrow 13.3\%}$} & 1.2304\textcolor{darkgreen}{$_{\downarrow 6.7\%}$} & 18.31\textcolor{darkred}{$_{\downarrow 48.9\%}$} & 81.5 \\
 \rowcolor{lightyellow} Stage 2 only & CLIP-L & $512^2$ & 304.20 & 262.19\textcolor{darkgreen}{$_{\downarrow 15.6\%}$} & 1.2163\textcolor{darkgreen}{$_{\downarrow 7.7\%}$} & 41.56\textcolor{darkgreen}{$_{\uparrow 16.1\%}$} & 83.4 \\
\rowcolor{lightyellow} ViT-AdaLA (Ours) & CLIP-L & $512^2$ & 304.20 & 262.19\textcolor{darkgreen}{$_{\downarrow 15.6\%}$} & 1.2163\textcolor{darkgreen}{$_{\downarrow 7.7\%}$} & 41.56\textcolor{darkgreen}{$_{\uparrow 16.1\%}$} & 85.5 \\
\bottomrule
\end{tabular}
}
\end{table*}
\FloatBarrier

\subsubsection{Experiments on Segmentation Tasks}
Tab. \ref{tab:ade20k_bench_VFMs} and Tab. \ref{tab:cityscapes_IN1K_bench} present the segmentation results for ADE20K and Cityscapes, respectively. From the results we can see that the overall performance of linear attention baselines have larger gaps with the softmax upper bound compared to the classification results. This is because segmentation requires stronger prior knowledge from the VFMs than the classification tasks do. Despite these challenges, ViT-AdaLA maintains competitive performance with softmax attention, demonstrating its superior ability to extract and preserve critical priors. Furthermore, ViT-AdaLA scales to high-resolution inputs with significant efficiency gains while suffering minimal performance degradation compared to the standard softmax attention.
\begin{table*}[!htbp]
\centering
\caption{mIoU fine-tuning comparison on Cityscapes under different input resolutions (512 vs. 1024) based on IN1K ViT-L and SigLIP-L. Native/direct conversions use the same downstream full fine-tuning budget; see Table~\ref{tab:baseline_protocol}. The segmentation head is Mask2former for all the methods.}
\label{tab:cityscapes_IN1K_bench}
\setlength{\tabcolsep}{4pt}
\renewcommand{\arraystretch}{1.1}
\scalebox{0.85}{
\begin{tabular}{p{3.3cm} c c c c c c c}
\toprule
& \multirow{2}{*}{\textbf{Backbone}} & \multirow{2}{*}{\textbf{Res.}} & \textbf{Params} & {\textbf{FLOPS}} & \textbf{Peak Mem.} & {\textbf{Throughput}} & \textbf{mIoU} \\
& & & \textbf{(M)} & \textbf{(G)} & \textbf{(GB)} & \textbf{(imgs/s)} & \textbf{(\%)} \\
\midrule
\rowcolor{gray!10} Softmax \cite{oquab2024dinov2} & IN1K ViT-L & $512^2$ & 304.15 & 310.60 & 1.3181 & 36.52 & 65.11 \\
Hedgehog \cite{zhanghedgehog}  & IN1K ViT-L & $512^2$ & 305.72 & 265.41\textcolor{darkgreen}{$_{\downarrow 14.5\%}$} & 1.2221\textcolor{darkgreen}{$_{\downarrow 7.3\%}$} & 37.44\textcolor{darkgreen}{$_{\uparrow 2.5\%}$}  & 37.69 \\
LoLCATS \cite{zhanglolcats}  & IN1K ViT-L & $512^2$ & 305.72 & 265.41\textcolor{darkgreen}{$_{\downarrow 14.5\%}$} & 1.2221\textcolor{darkgreen}{$_{\downarrow 7.3\%}$} & 37.44\textcolor{darkgreen}{$_{\uparrow 2.5\%}$} & 25.04 \\
Linformer \cite{wang2020linformer} & IN1K ViT-L & $512^2$ & 305.72 & 262.19\textcolor{darkgreen}{$_{\downarrow 15.6\%}$} & 1.2221\textcolor{darkgreen}{$_{\downarrow 7.3\%}$} & 45.41\textcolor{darkgreen}{$_{\uparrow 25.3\%}$} & 17.30 \\
Performer \cite{choromanskirethinking} & IN1K ViT-L & $512^2$ & 304.15 & 265.41\textcolor{darkgreen}{$_{\downarrow 14.6\%}$} & 1.2164\textcolor{darkgreen}{$_{\downarrow 7.7\%}$} & 36.48\textcolor{darkgreen}{$_{\uparrow 0.1\%}$} & 39.44 \\
Cosformer \cite{qincosformer} & IN1K ViT-L & $512^2$ & 304.15 & 265.41\textcolor{darkgreen}{$_{\downarrow 14.6\%}$} & 1.2226\textcolor{darkgreen}{$_{\downarrow 7.2\%}$} & 39.55\textcolor{darkgreen}{$_{\uparrow 8.2\%}$} & 39.33 \\
Nystr\"{o}mformer  \cite{xiong2021nystromformer} & IN1K ViT-L & $512^2$ & 304.15 & 265.20\textcolor{darkgreen}{$_{\downarrow 14.6\%}$} & 1.2163\textcolor{darkgreen}{$_{\downarrow 7.7\%}$} & 18.29\textcolor{darkred}{$_{\downarrow 50.8\%}$} & 45.58 \\
Monarch \cite{yaras2025monarchattention} & IN1K ViT-L & $512^2$ & 304.15 & 269.44\textcolor{darkgreen}{$_{\downarrow 13.3\%}$} & 1.2304\textcolor{darkgreen}{$_{\downarrow 6.7\%}$} & 18.31\textcolor{darkred}{$_{\downarrow 48.9\%}$} & 48.35 \\
 \rowcolor{lightyellow} Stage 2 only & IN1K ViT-L & $512^2$ & 304.15 & 262.19\textcolor{darkgreen}{$_{\downarrow 15.6\%}$} & 1.2163\textcolor{darkgreen}{$_{\downarrow 7.7\%}$} & 41.56\textcolor{darkgreen}{$_{\uparrow 16.1\%}$} & 62.24 \\
\rowcolor{lightyellow} ViT-AdaLA (Ours) & IN1K ViT-L & $512^2$ & 304.15 & 262.19\textcolor{darkgreen}{$_{\downarrow 15.6\%}$} & 1.2163\textcolor{darkgreen}{$_{\downarrow 7.7\%}$} & 41.56\textcolor{darkgreen}{$_{\uparrow 16.1\%}$} & 64.16 \\
\midrule
\rowcolor{gray!10} Softmax \cite{oquab2024dinov2} & IN1K ViT-L & $1024^2$ & 304.15 & 1241.0 & 3.2836 & 7.07 & 72.21 \\
Hedgehog \cite{zhanghedgehog}  & IN1K ViT-L & $1024^2$ & 305.72 & 442.24\textcolor{darkgreen}{$_{\downarrow 64.4\%}$} & 1.3823\textcolor{darkgreen}{$_{\downarrow 57.9\%}$} & 15.38\textcolor{darkgreen}{$_{\uparrow 117.4\%}$} & 38.07 \\
LoLCATS \cite{zhanglolcats}  & IN1K ViT-L & $1024^2$ & 305.72 & 442.24\textcolor{darkgreen}{$_{\downarrow 64.4\%}$} & 1.3823\textcolor{darkgreen}{$_{\downarrow 57.9\%}$} & 15.38\textcolor{darkgreen}{$_{\uparrow 117.4\%}$} & 25.98 \\
Linformer \cite{wang2020linformer} & IN1K ViT-L & $1024^2$ & 305.72 & 429.35\textcolor{darkgreen}{$_{\downarrow 65.4\%}$} & 1.3999\textcolor{darkgreen}{$_{\downarrow 57.4\%}$} & 17.83\textcolor{darkgreen}{$_{\uparrow 152.0\%}$} & 7.78 \\
Performer \cite{choromanskirethinking} & IN1K ViT-L & $1024^2$ & 304.15 & 442.24\textcolor{darkgreen}{$_{\downarrow 64.4\%}$} & 1.3765\textcolor{darkgreen}{$_{\downarrow 58.1\%}$} & 15.23\textcolor{darkgreen}{$_{\uparrow 115.3\%}$} & 40.86 \\
Cosformer \cite{qincosformer} & IN1K ViT-L & $1024^2$ & 304.15 & 442.24\textcolor{darkgreen}{$_{\downarrow 64.4\%}$} & 1.3945\textcolor{darkgreen}{$_{\downarrow 57.5\%}$} & 15.29\textcolor{darkgreen}{$_{\uparrow 116.1\%}$} & 40.14 \\
Nystr\"{o}mformer  \cite{xiong2021nystromformer} & IN1K ViT-L & $1024^2$ & 304.15 & 442.24\textcolor{darkgreen}{$_{\downarrow 64.4\%}$} & 1.3764\textcolor{darkgreen}{$_{\downarrow 58.1\%}$} & 14.61\textcolor{darkgreen}{$_{\uparrow 106.5\%}$} & 43.15 \\
Monarch \cite{yaras2025monarchattention} & IN1K ViT-L & $1024^2$ & 304.15 & 500.24\textcolor{darkgreen}{$_{\downarrow 59.7\%}$} & 1.5469\textcolor{darkgreen}{$_{\downarrow 52.9\%}$} & 9.74\textcolor{darkgreen}{$_{\uparrow 37.6\%}$} & 52.18 \\
\rowcolor{lightyellow} Stage 2 only & IN1K ViT-L & $1024^2$ & 304.15 & 429.35\textcolor{darkgreen}{$_{\downarrow 65.4\%}$} & 1.3764\textcolor{darkgreen}{$_{\downarrow 58.1\%}$} & 15.95\textcolor{darkgreen}{$_{\uparrow 125.4\%}$} & 66.42 \\
\rowcolor{lightyellow} ViT-AdaLA (Ours) & IN1K ViT-L & $1024^2$ & 304.15 & 429.35\textcolor{darkgreen}{$_{\downarrow 65.4\%}$} & 1.3764\textcolor{darkgreen}{$_{\downarrow 58.1\%}$} & 15.95\textcolor{darkgreen}{$_{\uparrow 125.4\%}$} & 68.25 \\
\midrule
\rowcolor{gray!10} Softmax \cite{oquab2024dinov2} & SigLIP-L & $512^2$ & 304.15 & 310.60 & 1.3181 & 36.52 & 70.57 \\
Hedgehog \cite{zhanghedgehog}  & SigLIP-L & $512^2$ & 305.72 & 265.41\textcolor{darkgreen}{$_{\downarrow 14.5\%}$} & 1.2221\textcolor{darkgreen}{$_{\downarrow 7.3\%}$} & 37.44\textcolor{darkgreen}{$_{\uparrow 2.5\%}$}  & 39.53 \\
LoLCATS \cite{zhanglolcats}  & SigLIP-L & $512^2$ & 305.72 & 265.41\textcolor{darkgreen}{$_{\downarrow 14.5\%}$} & 1.2221\textcolor{darkgreen}{$_{\downarrow 7.3\%}$} & 37.44\textcolor{darkgreen}{$_{\uparrow 2.5\%}$} & 23.09 \\
Linformer \cite{wang2020linformer} & SigLIP-L & $512^2$ & 305.72 & 262.19\textcolor{darkgreen}{$_{\downarrow 15.6\%}$} & 1.2221\textcolor{darkgreen}{$_{\downarrow 7.3\%}$} & 45.41\textcolor{darkgreen}{$_{\uparrow 25.3\%}$} & 21.81 \\
Performer \cite{choromanskirethinking} & SigLIP-L & $512^2$ & 304.15 & 265.41\textcolor{darkgreen}{$_{\downarrow 14.6\%}$} & 1.2164\textcolor{darkgreen}{$_{\downarrow 7.7\%}$} & 36.48\textcolor{darkgreen}{$_{\uparrow 0.1\%}$} & 39.93 \\
Cosformer \cite{qincosformer} & SigLIP-L & $512^2$ & 304.15 & 265.41\textcolor{darkgreen}{$_{\downarrow 14.6\%}$} & 1.2226\textcolor{darkgreen}{$_{\downarrow 7.2\%}$} & 39.55\textcolor{darkgreen}{$_{\uparrow 8.2\%}$} & 40.56 \\
Nystr\"{o}mformer  \cite{xiong2021nystromformer} & SigLIP-L & $512^2$ & 304.15 & 265.20\textcolor{darkgreen}{$_{\downarrow 14.6\%}$} & 1.2163\textcolor{darkgreen}{$_{\downarrow 7.7\%}$} & 18.29\textcolor{darkred}{$_{\downarrow 50.8\%}$} & 43.52 \\
Monarch \cite{yaras2025monarchattention} & SigLIP-L & $512^2$ & 304.15 & 269.44\textcolor{darkgreen}{$_{\downarrow 13.3\%}$} & 1.2304\textcolor{darkgreen}{$_{\downarrow 6.7\%}$} & 18.31\textcolor{darkred}{$_{\downarrow 48.9\%}$} & 43.94 \\
 \rowcolor{lightyellow} Stage 2 only & SigLIP-L & $512^2$ & 304.15 & 262.19\textcolor{darkgreen}{$_{\downarrow 15.6\%}$} & 1.2163\textcolor{darkgreen}{$_{\downarrow 7.7\%}$} & 41.56\textcolor{darkgreen}{$_{\uparrow 16.1\%}$} & 67.35 \\
\rowcolor{lightyellow} ViT-AdaLA (Ours) & SigLIP-L & $512^2$ & 304.15 & 262.19\textcolor{darkgreen}{$_{\downarrow 15.6\%}$} & 1.2163\textcolor{darkgreen}{$_{\downarrow 7.7\%}$} & 41.56\textcolor{darkgreen}{$_{\uparrow 16.1\%}$} & 68.20 \\
\midrule
\rowcolor{gray!10} Softmax \cite{oquab2024dinov2} & SigLIP-L & $1024^2$ & 304.15 & 1241.0 & 3.2836 & 7.07 & 76.53 \\
Hedgehog \cite{zhanghedgehog}  & SigLIP-L & $1024^2$ & 305.72 & 442.24\textcolor{darkgreen}{$_{\downarrow 64.4\%}$} & 1.3823\textcolor{darkgreen}{$_{\downarrow 57.9\%}$} & 15.38\textcolor{darkgreen}{$_{\uparrow 117.4\%}$} & 40.17 \\
LoLCATS \cite{zhanglolcats}  & SigLIP-L & $1024^2$ & 305.72 & 442.24\textcolor{darkgreen}{$_{\downarrow 64.4\%}$} & 1.3823\textcolor{darkgreen}{$_{\downarrow 57.9\%}$} & 15.38\textcolor{darkgreen}{$_{\uparrow 117.4\%}$} & 24.56 \\
Linformer \cite{wang2020linformer} & SigLIP-L & $1024^2$ & 305.72 & 429.35\textcolor{darkgreen}{$_{\downarrow 65.4\%}$} & 1.3999\textcolor{darkgreen}{$_{\downarrow 57.4\%}$} & 17.83\textcolor{darkgreen}{$_{\uparrow 152.0\%}$} & 6.92 \\
Performer \cite{choromanskirethinking} & SigLIP-L & $1024^2$ & 304.15 & 442.24\textcolor{darkgreen}{$_{\downarrow 64.4\%}$} & 1.3765\textcolor{darkgreen}{$_{\downarrow 58.1\%}$} & 15.23\textcolor{darkgreen}{$_{\uparrow 115.3\%}$} & 40.53 \\
Cosformer \cite{qincosformer} & SigLIP-L & $1024^2$ & 304.15 & 442.24\textcolor{darkgreen}{$_{\downarrow 64.4\%}$} & 1.3945\textcolor{darkgreen}{$_{\downarrow 57.5\%}$} & 15.29\textcolor{darkgreen}{$_{\uparrow 116.1\%}$} & 39.90 \\
Nystr\"{o}mformer  \cite{xiong2021nystromformer} & SigLIP-L & $1024^2$ & 304.15 & 442.24\textcolor{darkgreen}{$_{\downarrow 64.4\%}$} & 1.3764\textcolor{darkgreen}{$_{\downarrow 58.1\%}$} & 14.61\textcolor{darkgreen}{$_{\uparrow 106.5\%}$} & 41.86 \\
Monarch \cite{yaras2025monarchattention} & SigLIP-L & $1024^2$ & 304.15 & 500.24\textcolor{darkgreen}{$_{\downarrow 59.7\%}$} & 1.5469\textcolor{darkgreen}{$_{\downarrow 52.9\%}$} & 9.74\textcolor{darkgreen}{$_{\uparrow 37.6\%}$} & 46.58 \\
\rowcolor{lightyellow} Stage 2 only & SigLIP-L & $1024^2$ & 304.15 & 429.35\textcolor{darkgreen}{$_{\downarrow 65.4\%}$} & 1.3764\textcolor{darkgreen}{$_{\downarrow 58.1\%}$} & 15.95\textcolor{darkgreen}{$_{\uparrow 125.4\%}$} & 72.41 \\
\rowcolor{lightyellow} ViT-AdaLA (Ours) & SigLIP-L & $1024^2$ & 304.15 & 429.35\textcolor{darkgreen}{$_{\downarrow 65.4\%}$} & 1.3764\textcolor{darkgreen}{$_{\downarrow 58.1\%}$} & 15.95\textcolor{darkgreen}{$_{\uparrow 125.4\%}$} & 73.33 \\
\bottomrule
\end{tabular}
}
\end{table*}
\begin{table*}[!htbp]
\centering
\caption{mIoU fine-tuning comparison on ADE20K under ViT-IN1K and SigLIP with multiple linear attention baselines. Native/direct conversions use the same downstream full fine-tuning budget; see Table~\ref{tab:baseline_protocol}. The segmentation head is Mask2former for all the methods.}
\label{tab:ade20k_bench_VFMs}
\setlength{\tabcolsep}{4pt}
\renewcommand{\arraystretch}{1.1}
\scalebox{0.85}{
\begin{tabular}{p{3.5cm} c c c c c c c}
\toprule
& \multirow{2}{*}{\textbf{Backbone}} & \multirow{2}{*}{\textbf{Res.}} & \textbf{Params} & {\textbf{FLOPS}} & \textbf{Peak Mem.} & {\textbf{Throughput}} & \textbf{mIoU} \\
& & & \textbf{(M)} & \textbf{(G)} & \textbf{(GB)} & \textbf{(imgs/s)} & \textbf{(\%)} \\
\midrule
\rowcolor{gray!10} Softmax \cite{oquab2024dinov2} & IN1K ViT-L & $512^2$ & 304.20 & 310.60 & 1.3181 & 36.52 & 50.83 \\
Hedgehog \cite{zhanghedgehog}  & IN1K ViT-L & $512^2$ & 305.77 & 265.41\textcolor{darkgreen}{$_{\downarrow 14.5\%}$} & 1.2221\textcolor{darkgreen}{$_{\downarrow 7.3\%}$} & 37.44\textcolor{darkgreen}{$_{\uparrow 2.5\%}$}  & 25.15 \\
LoLCATS \cite{zhanglolcats}  & IN1K ViT-L & $512^2$ & 305.77 & 265.41\textcolor{darkgreen}{$_{\downarrow 14.5\%}$} & 1.2221\textcolor{darkgreen}{$_{\downarrow 7.3\%}$} & 37.44\textcolor{darkgreen}{$_{\uparrow 2.5\%}$} & 20.17 \\
Linformer \cite{wang2020linformer} & IN1K ViT-L & $512^2$ & 305.77 & 262.19\textcolor{darkgreen}{$_{\downarrow 15.6\%}$} & 1.2221\textcolor{darkgreen}{$_{\downarrow 7.3\%}$} & 45.41\textcolor{darkgreen}{$_{\uparrow 25.3\%}$} & 11.49 \\
Performer \cite{choromanskirethinking} & IN1K ViT-L & $512^2$ & 304.20 & 265.41\textcolor{darkgreen}{$_{\downarrow 14.6\%}$} & 1.2164\textcolor{darkgreen}{$_{\downarrow 7.7\%}$} & 36.48\textcolor{darkgreen}{$_{\uparrow 0.1\%}$} & 31.01 \\
Cosformer \cite{qincosformer} & IN1K ViT-L & $512^2$ & 304.20 & 265.41\textcolor{darkgreen}{$_{\downarrow 14.6\%}$} & 1.2226\textcolor{darkgreen}{$_{\downarrow 7.2\%}$} & 39.55\textcolor{darkgreen}{$_{\uparrow 8.2\%}$} & 29.65 \\
Nystr\"{o}mformer  \cite{xiong2021nystromformer} & IN1K ViT-L & $512^2$ & 304.20 & 265.20\textcolor{darkgreen}{$_{\downarrow 14.6\%}$} & 1.2163\textcolor{darkgreen}{$_{\downarrow 7.7\%}$} & 18.29\textcolor{darkred}{$_{\downarrow 50.8\%}$} & 41.67 \\
Monarch \cite{yaras2025monarchattention} & IN1K ViT-L & $512^2$ & 304.20 & 269.44\textcolor{darkgreen}{$_{\downarrow 13.3\%}$} & 1.2304\textcolor{darkgreen}{$_{\downarrow 6.7\%}$} & 18.31\textcolor{darkred}{$_{\downarrow 48.9\%}$} & 45.05 \\
 \rowcolor{lightyellow} Stage 2 only & IN1K ViT-L & $512^2$ & 304.20 & 262.19\textcolor{darkgreen}{$_{\downarrow 15.6\%}$} & 1.2163\textcolor{darkgreen}{$_{\downarrow 7.7\%}$} & 41.56\textcolor{darkgreen}{$_{\uparrow 16.1\%}$} & 46.92 \\
\rowcolor{lightyellow} ViT-AdaLA (Ours) & IN1K ViT-L & $512^2$ & 304.20 & 262.19\textcolor{darkgreen}{$_{\downarrow 15.6\%}$} & 1.2163\textcolor{darkgreen}{$_{\downarrow 7.7\%}$} & 41.56\textcolor{darkgreen}{$_{\uparrow 16.1\%}$} & 49.75 \\
\midrule
\rowcolor{gray!10} Softmax \cite{oquab2024dinov2} & SigLIP-L & $512^2$ & 304.20 & 310.60 & 1.3181 & 36.52 & 54.4 \\
Hedgehog \cite{zhanghedgehog}  & SigLIP-L & $512^2$ & 305.77 & 265.41\textcolor{darkgreen}{$_{\downarrow 14.5\%}$} & 1.2221\textcolor{darkgreen}{$_{\downarrow 7.3\%}$} & 37.44\textcolor{darkgreen}{$_{\uparrow 2.5\%}$}  & 22.13 \\
LoLCATS \cite{zhanglolcats}  & SigLIP-L & $512^2$ & 305.77 & 265.41\textcolor{darkgreen}{$_{\downarrow 14.5\%}$} & 1.2221\textcolor{darkgreen}{$_{\downarrow 7.3\%}$} & 37.44\textcolor{darkgreen}{$_{\uparrow 2.5\%}$} & 20.11 \\
Linformer \cite{wang2020linformer} & SigLIP-L & $512^2$ & 305.77 & 262.19\textcolor{darkgreen}{$_{\downarrow 15.6\%}$} & 1.2221\textcolor{darkgreen}{$_{\downarrow 7.3\%}$} & 45.41\textcolor{darkgreen}{$_{\uparrow 25.3\%}$} & 12.37 \\
Performer \cite{choromanskirethinking} & SigLIP-L & $512^2$ & 304.20 & 265.41\textcolor{darkgreen}{$_{\downarrow 14.6\%}$} & 1.2164\textcolor{darkgreen}{$_{\downarrow 7.7\%}$} & 36.48\textcolor{darkgreen}{$_{\uparrow 0.1\%}$} & 23.97 \\
Cosformer \cite{qincosformer} & SigLIP-L & $512^2$ & 304.20 & 265.41\textcolor{darkgreen}{$_{\downarrow 14.6\%}$} & 1.2226\textcolor{darkgreen}{$_{\downarrow 7.2\%}$} & 39.55\textcolor{darkgreen}{$_{\uparrow 8.2\%}$} & 23.08 \\
Nystr\"{o}mformer  \cite{xiong2021nystromformer} & SigLIP-L & $512^2$ & 304.20 & 265.20\textcolor{darkgreen}{$_{\downarrow 14.6\%}$} & 1.2163\textcolor{darkgreen}{$_{\downarrow 7.7\%}$} & 18.29\textcolor{darkred}{$_{\downarrow 50.8\%}$} & 32.54 \\
Monarch \cite{yaras2025monarchattention} & SigLIP-L & $512^2$ & 304.20 & 269.44\textcolor{darkgreen}{$_{\downarrow 13.3\%}$} & 1.2304\textcolor{darkgreen}{$_{\downarrow 6.7\%}$} & 18.31\textcolor{darkred}{$_{\downarrow 48.9\%}$} & 38.87 \\
 \rowcolor{lightyellow} Stage 2 only & SigLIP-L & $512^2$ & 304.20 & 262.19\textcolor{darkgreen}{$_{\downarrow 15.6\%}$} & 1.2163\textcolor{darkgreen}{$_{\downarrow 7.7\%}$} & 41.56\textcolor{darkgreen}{$_{\uparrow 16.1\%}$} & 51.39 \\
\rowcolor{lightyellow} ViT-AdaLA (Ours) & SigLIP-L & $512^2$ & 304.20 & 262.19\textcolor{darkgreen}{$_{\downarrow 15.6\%}$} & 1.2163\textcolor{darkgreen}{$_{\downarrow 7.7\%}$} & 41.56\textcolor{darkgreen}{$_{\uparrow 16.1\%}$} & 53.55 \\
\bottomrule
\end{tabular}
}
\end{table*}

\FloatBarrier
\subsubsection{DINOv2-L efficiency performance with different batch sizes}
To better illustrate efficiency in standard ViT settings, where token sequence lengths are relatively small, we conducted a detailed analysis in Tab. \ref{tab:dinov2l_efficiency_batch_size} by scaling the batch size (BS) from 1, the default batch size in our main paper, to 128 using a DINOv2-L backbone.

\begin{table}[!htbp]
\centering
\caption{DINOv2-L efficiency performance with different batch sizes. The input size is $256 \times 256$. Relative changes are reported with respect to Softmax attention under the same batch size.}
\label{tab:dinov2l_efficiency_batch_size}
\resizebox{\linewidth}{!}{
\begin{tabular}{c l c c c}
\toprule
\textbf{Batch Size} & \textbf{Attention Type} & \textbf{Latency (ms)} & \textbf{Throughput (imgs/s)} & \textbf{Memory (GB)} \\
\midrule
\multirow{2}{*}{1}
& Softmax & 15.916 & 62.83 & 1.1770 \\
& Linear  & 20.291 {\small $(+27.5\%)$} & 49.28 {\small $(-21.6\%)$} & 1.1750 {\small $(-0.2\%)$} \\
\midrule
\multirow{2}{*}{16}
& Softmax & 63.420 & 252.29 & 1.4100 \\
& Linear  & 61.410 {\small $(-3.2\%)$} & 260.54 {\small $(+3.3\%)$} & 1.3625 {\small $(-3.4\%)$} \\
\midrule
\multirow{2}{*}{32}
& Softmax & 126.091 & 253.78 & 1.6595 \\
& Linear  & 121.086 {\small $(-4.0\%)$} & 264.28 {\small $(+4.1\%)$} & 1.5624 {\small $(-5.8\%)$} \\
\midrule
\multirow{2}{*}{64}
& Softmax & 257.052 & 248.98 & 2.1535 \\
& Linear  & 246.232 {\small $(-4.2\%)$} & 259.92 {\small $(+4.4\%)$} & 1.9653 {\small $(-8.7\%)$} \\
\midrule
\multirow{2}{*}{128}
& Softmax & 510.024 & 250.97 & 3.1430 \\
& Linear  & 487.622 {\small $(-4.4\%)$} & 262.50 {\small $(+4.6\%)$} & 2.7621 {\small $(-12.1\%)$} \\
\bottomrule
\end{tabular}
}
\end{table}

\FloatBarrier
\paragraph{Key Findings.}
At batch size one, linear attention is slower than softmax, but it gains throughput as the workload grows. At batch size 128, it is 4.6\% faster and uses 12.1\% less memory, consistent with lower attention-map memory pressure at larger batches.

\clearpage
\subsection{Protocols and Transfer Validation}
We clarify matched evaluation budgets and test transfer beyond the primary backbone and task.

\subsubsection{Baseline Protocol and Adaptation Cost}\label{sec:protocol_and_cost}
The main baseline tables compare complete conversion procedures. Native attention baselines do not receive ViT-AdaLA's feature-alignment stage; every method uses the same downstream data, head, and full fine-tuning schedule within each task. The Stage-2-only ablation starts from direct linear replacement. Table~\ref{tab:baseline_protocol} clarifies these protocols, and Table~\ref{tab:matched_attention_alignment} reports matched alignment experiments on alternative modules.
\begin{table}[!htbp]
\centering
\caption{Protocol for the main comparison tables. Native denotes each baseline's own conversion or local attention transfer, not ViT-AdaLA Stage 1.}
\label{tab:baseline_protocol}
\small
\resizebox{\linewidth}{!}{%
\begin{tabular}{lccc}
\toprule
Method group & Local conversion & ViT-AdaLA Stage 2 & Same downstream budget \\
\midrule
Softmax teacher & None & No & Yes \\
Hedgehog, LoLCATS & Native attention transfer & No & Yes \\
Linformer, Performer, Cosformer & Native/direct replacement & No & Yes \\
Nystr\"omformer, Monarch & Native/direct replacement & No & Yes \\
Vanilla linear, Stage 2 only & Direct replacement & Yes & Yes \\
ViT-AdaLA & Stage 1 & Yes & Yes \\
Nystr\"omformer, Monarch + our alignment & Stage 1 & Yes & Yes \\
\bottomrule
\end{tabular}%
}
\end{table}
\begin{table*}[!htbp]
\centering
\caption{Applying the same Stage 1 attention alignment and Stage 2 feature alignment to different DINOv2-L attention modules. Native/direct variants use their original conversion procedures; all rows use the same downstream full fine-tuning budget.}
\label{tab:matched_attention_alignment}
\small
\setlength{\tabcolsep}{5pt}
\begin{tabular}{lccc}
\toprule
Attention module & Stages 1+2 & IN-1K top-1 (\%) & ADE20K mIoU \\
\midrule
Nystr\"omformer (native) & No & 82.4 & 41.16 \\
Nystr\"omformer + alignment & Yes & 85.3 & 54.60 \\
Monarch (native) & No & 82.7 & 44.95 \\
Monarch + alignment & Yes & 85.7 & 54.90 \\
Vanilla + alignment & Yes & \textbf{86.0} & \textbf{55.55} \\
\bottomrule
\end{tabular}
\end{table*}

Stage 2 uses ImageNet-22K and is a substantial one-time adaptation step. Table~\ref{tab:stage_compute} converts the measured per-epoch wall times in Table~\ref{tab:training_time} into aggregate H100 GPU-hours. These figures cover Stages 1 and 2 only; downstream Stage 3 compute is task-dependent and is excluded.
\begin{table}[!htbp]
\centering
\caption{Measured adaptation cost on eight H100 GPUs, excluding downstream Stage 3. GPU-hours equal eight times wall-clock hours.}
\label{tab:stage_compute}
\small
\resizebox{\linewidth}{!}{%
\begin{tabular}{lcccc}
\toprule
Backbone & Stage 1 epochs & Stage 1 GPU-h & Stage 2 epochs & Stage 2 GPU-h \\
\midrule
DINOv2-L & 4 & 2.46 & 30 & 2595.1 \\
SigLIP-L & 4 & 2.30 & 10 & 909.7 \\
CLIP-L & 4 & 3.16 & 15 & 1338.7 \\
IN-1K ViT-L & 4 & 2.44 & 20 & 1607.7 \\
\bottomrule
\end{tabular}%
}
\end{table}

\FloatBarrier
\subsubsection{Held-Out Feature Organization}\label{sec:heldout_geometry}
Frozen-backbone transfer is reported in Table~\ref{tab:frozen_transfer}; here we examine whether the teacher's feature-space structure is preserved without training a task head.
We evaluate the DINOv2-L teacher, direct linear replacement, and the Stage-2 ViT-AdaLA checkpoint before downstream fine-tuning. The complete Flowers-102 test split (6,149 images) was unused in Stages 1--2 and hyperparameter selection. All models receive identical $512^2$ inputs and aligned patch tokens; no head or model parameter is trained. Nearest neighbors use mean-pooled final patch features with $k=20$.
\begin{table}[!htbp]
\centering
\caption{Feature-space agreement with the frozen softmax DINOv2-L teacher on held-out Flowers-102. CKA and nearest-neighbor overlap use pooled final features; cosine similarity uses aligned tokens.}
\label{tab:heldout_geometry}
\small
\resizebox{\linewidth}{!}{%
\begin{tabular}{lcccc}
\toprule
Linearized model & CKA & NN overlap@20 & Token cosine & Class-centroid corr. \\
\midrule
Direct replacement & 0.196 & 10.0\% & 0.014 & 0.098 \\
ViT-AdaLA after Stage 2 & \textbf{0.993} & \textbf{90.5\%} & \textbf{0.940} & \textbf{0.997} \\
\bottomrule
\end{tabular}%
}
\end{table}
The metrics test global geometry, sample neighborhoods, spatial token correspondence, and class-cluster structure. High agreement does not imply exact preservation on every dataset; it supports the frozen-head transfer in Table~\ref{tab:frozen_transfer}.

\FloatBarrier
\subsection{Alignment Objectives and Teacher Ablations}
We detail checkpoint selection and test alternative parameterizations, joint training, and teacher choices.

\subsubsection{Alignment Objective and Checkpoint Selection}\label{sec:alignment_objective_details}
The Stage-1 objective sums per-block mean-squared errors rather than averaging over the number of blocks; each student attention module receives the corresponding frozen teacher input. Stage 2 instead updates the complete student to align its final features. The per-backbone epoch budgets and feature-loss weights $\lambda$ are listed in Table~\ref{tab:training_stage_configuration}; no separate early-stopping criterion is specified by the reported curves.

\subsubsection{LoLCATS Parameterization and Joint-Stage Ablations}\label{sec:additional_stage_ablations}
We compare LoLCATS with LoRA and with full-weight tuning under the same DINOv2-L downstream evaluation. Both omit task-agnostic Stage-2 feature alignment. Increased tuning capacity raises ImageNet-1K accuracy by 0.7 points and ADE20K mIoU by 0.40, far less than the gain obtained with global feature alignment.
\begin{table}[!htbp]
\centering
\caption{LoLCATS tuning capacity versus the complete ViT-AdaLA pipeline on DINOv2-L. All rows use the same downstream full fine-tuning budget.}
\label{tab:lolcats_full_weights}
\small
\begin{tabular}{lccc}
\toprule
Method & Feature alignment & IN-1K top-1 & ADE20K mIoU \\
\midrule
LoLCATS, LoRA & No & 61.6 & 17.42 \\
LoLCATS, full weights & No & 62.3 & 17.82 \\
ViT-AdaLA & Yes & \textbf{86.0} & \textbf{55.55} \\
\bottomrule
\end{tabular}
\end{table}
Stage 1 compares softmax and linear attention on identical teacher inputs, whereas Stage 2 changes student intermediate features end to end. We tested a simultaneous proxy that relaxes the identical-input condition and combines the losses. On ADE20K with SigLIP-L, it offers almost no improvement over Stage 2 alone; the sequential pipeline performs better (Table~\ref{tab:joint_alignment}).
\begin{table}[!htbp]
\centering
\caption{Sequential versus simultaneous alignment on ADE20K with SigLIP-L. The joint proxy relaxes the matched-input condition of Stage 1.}
\label{tab:joint_alignment}
\small
\begin{tabular}{lc}
\toprule
Training strategy & ADE20K mIoU \\
\midrule
Stage 2 only & 51.39 \\
Simultaneous Stage 1 + Stage 2 & 51.42 \\
Sequential Stage 1 $\rightarrow$ Stage 2 & \textbf{53.16} \\
\bottomrule
\end{tabular}
\end{table}

\FloatBarrier

\FloatBarrier
\section{More Visualization Results}\label{sec:vis_results}
\begin{figure}
    \centering
    \includegraphics[width=1\linewidth]{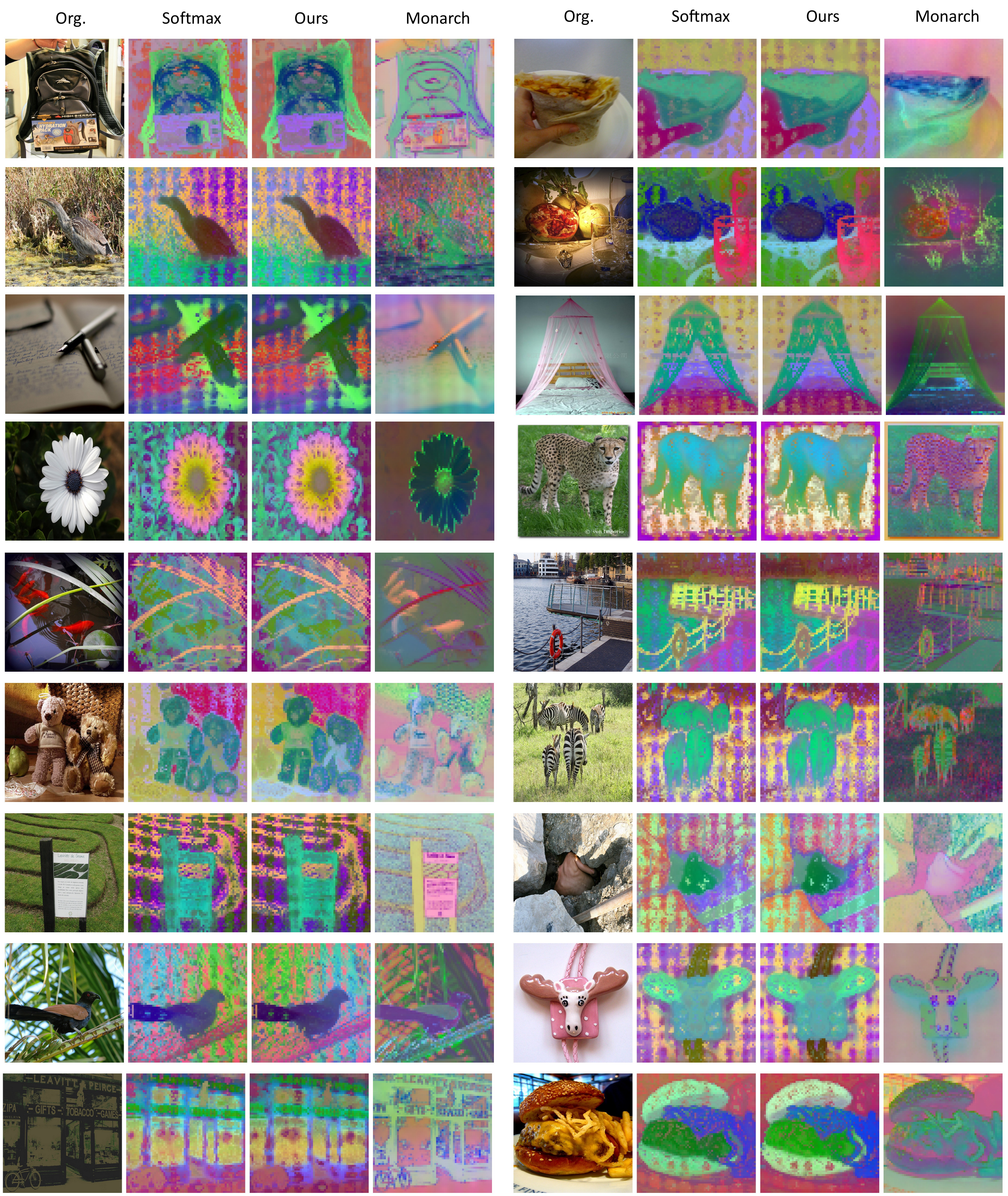}
    \caption{Visualization of PCA-projected features from the final layer of DINOv2-L. Original softmax features, ViT-AdaLA (ours) and Monarch Attention features are compared by projecting to three channels using PCA. These results indicate that ViT-AdaLA can learn more prior knowledge from the original VFM.}
    \label{fig:pca_monarch_more_imgs}
\end{figure}

\begin{figure}
    \centering
    \includegraphics[width=0.99\linewidth]{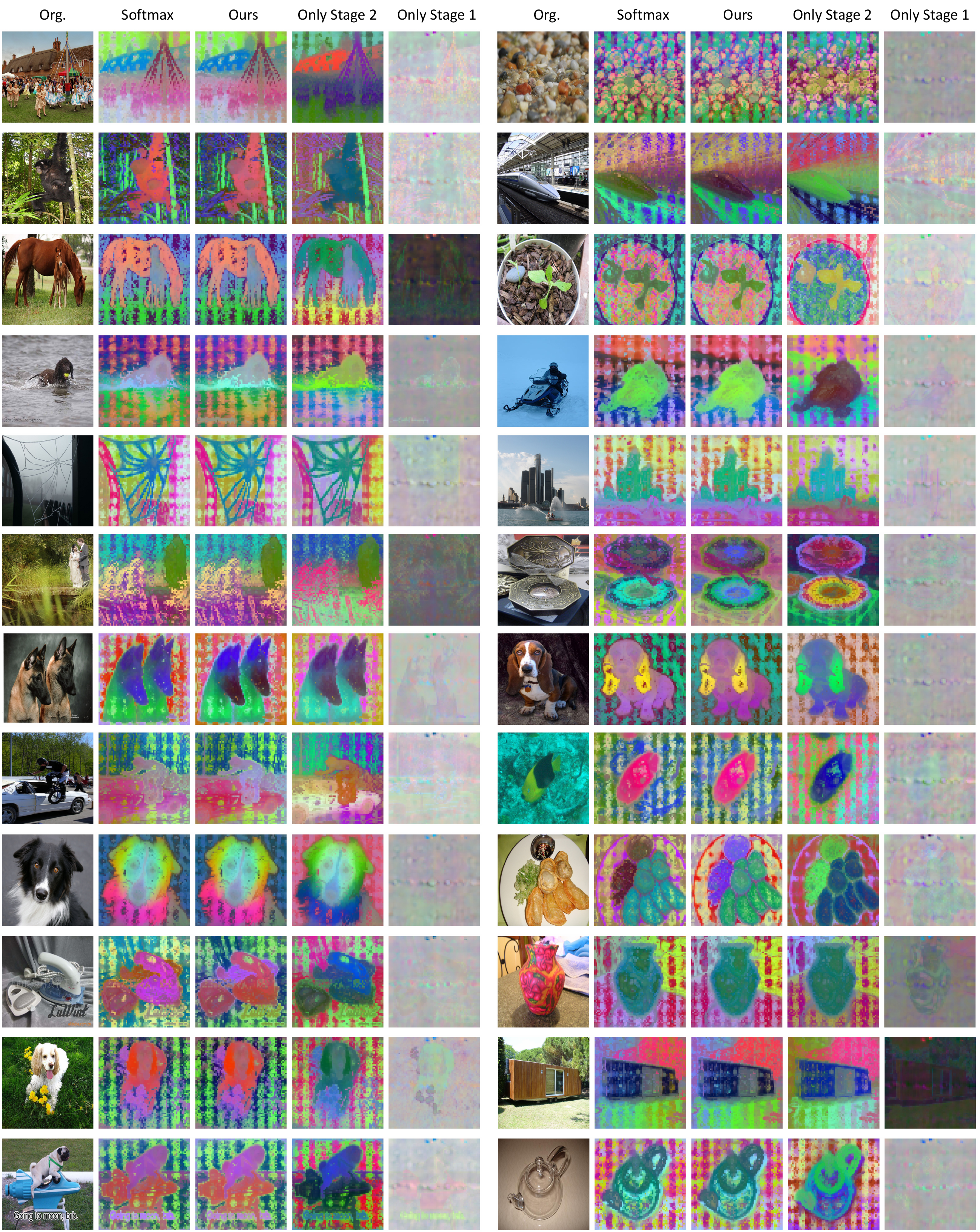}
    \caption{Visualization of PCA-projected features from the final layer of DINOv2-L. We ablate the Stages 1 and 2 training procedures and visualize the resulting PCA features for comparison. The results indicate that Stage 2 is crucial for extracting prior knowledge from VFMs, while the inclusion of Stage 1 further enhances this knowledge transfer.}
    \label{fig:pca_ablation}
\end{figure}

Additional PCA visualizations are provided in Figures \ref{fig:pca_monarch_more_imgs} and \ref{fig:pca_ablation}. Specifically, Figure \ref{fig:pca_monarch_more_imgs} presents a comparative analysis with Monarch Attention, while Figure \ref{fig:pca_ablation} illustrates the impact of Stage 1 and Stage 2 through ablation-based PCA results. 

Fig. \ref{fig:pca_monarch_more_imgs} shows that ViT-AdaLA more closely matches the prior features of VFMs than approximation-based linear attention methods such as Monarch, demonstrating its superior ability to distill prior knowledge from softmax-based VFMs. Fig. \ref{fig:pca_ablation} analyzes the contribution of different stages, showing that Stage 2 effectively preserves most of the original VFM features, while combining both stages yields the strongest retention of prior knowledge.

\FloatBarrier
\section{Limitations}\label{sec:limitations}
ViT-AdaLA has several limitations. Its strongest measured inference gains arise in high-resolution dense prediction: at $256^2$ and batch size one it is slower than softmax (Table~\ref{tab:resolution_crossover}). Stage 2 requires ImageNet-22K and substantial one-time H100 compute (Table~\ref{tab:stage_compute}); the scratch reference does not match its data or supervision, and complete totals also depend on task-specific Stage 3 fine-tuning. We evaluate classification, semantic segmentation, and one foreground-segmentation setting, leaving detection and generation untested. We do not report multi-seed uncertainty for the new ablations. Frozen-head transfer preserves most of the teacher's performance on the reported tasks, but this does not establish exact preservation of every feature or task. Improving adaptation efficiency and testing more tasks and backbones remain future work.

\section{Future Directions}
In the future, this linear attention can be extended to vision large language models (VLLMs \cite{li2025visual}), which process long visual sequences and thus incur substantial computational overhead. By replacing quadratic self-attention with linear variants, these models could achieve significantly improved scalability when handling high-resolution images or long video inputs.
Beyond efficiency, this direction also opens up opportunities for better interpretability. For example, integrating explanation methods \cite{chenless, chen2025mllms, chen2025interpreting} with linear attention could provide more transparent insights into how visual tokens contribute to model predictions. This is particularly valuable in multimodal settings, where understanding cross-modal interactions remains challenging.

\section{Impact Statement}
Our work focuses on adapting pretrained ViTs to linear attention while preserving their reusable representations. Its strongest measured inference benefits occur for long visual sequences, which could make high-resolution perception more practical. The one-time adaptation stage still requires substantial compute and should be included when assessing deployment costs.

From an ethical perspective, the main experiments use public datasets, while the SAM-H foreground-segmentation ablation uses an internal training set and SAM-HQ for evaluation. The method inherits risks associated with foundation models, including pretraining-data bias and misuse in privacy-sensitive applications. Efficient deployment can broaden both benefits and risks; dataset auditing and responsible use remain important.

Looking forward, we hope our work encourages further research into efficient and adaptable vision models that balance performance, scalability, and responsible deployment. We do not foresee any immediate negative societal impacts arising uniquely from this work beyond those already associated with large-scale visual recognition systems.

\stopcontents[appendix]
\section*{NeurIPS Paper Checklist}

IMPORTANT, please:
\begin{itemize}
    \item {\bf Delete this instruction block, but keep the section heading ``NeurIPS Paper Checklist"},
    \item  {\bf Keep the checklist subsection headings, questions/answers and guidelines below.}
    \item {\bf Do not modify the questions and only use the provided macros for your answers}.
\end{itemize}


\begin{enumerate}

\item {\bf Claims}
    \item[] Question: Do the main claims made in the abstract and introduction accurately reflect the paper's contributions and scope?
    \item[] Answer: \answerYes{} 
    \item[] Justification: The main claims in the abstract and introduction accurately reflect the paper's contributions and scope.

\item {\bf Limitations}
    \item[] Question: Does the paper discuss the limitations of the work performed by the authors?
    \item[] Answer: \answerYes{} 
    \item[] Justification: 
    Limitations can be found the App. \ref{sec:limitations}.

\item {\bf Theory assumptions and proofs}
    \item[] Question: For each theoretical result, does the paper provide the full set of assumptions and a complete (and correct) proof?
    \item[] Answer: \answerNA{} 
    \item[] Justification: 
    This paper doesn't include theoretical results.

    \item {\bf Experimental result reproducibility}
    \item[] Question: Does the paper fully disclose all the information needed to reproduce the main experimental results of the paper to the extent that it affects the main claims and/or conclusions of the paper (regardless of whether the code and data are provided or not)?
    \item[] Answer: \answerYes{} 
    \item[] Justification: 
    The paper fully disclose all the information to reproduce the main experiment results in the main content and appendix.

\item {\bf Open access to data and code}
    \item[] Question: Does the paper provide open access to the data and code, with sufficient instructions to faithfully reproduce the main experimental results, as described in supplemental material?
    \item[] Answer: \answerNo{} 
    \item[] Justification: 
    We would like to open source our codes and data once our paper is accepted.

\item {\bf Experimental setting/details}
    \item[] Question: Does the paper specify all the training and test details (e.g., data splits, hyperparameters, how they were chosen, type of optimizer) necessary to understand the results?
    \item[] Answer: \answerYes{} 
    \item[] Justification:
    We have specify all the training and test details necessary to understand the results.

\item {\bf Experiment statistical significance}
    \item[] Question: Does the paper report error bars suitably and correctly defined or other appropriate information about the statistical significance of the experiments?
    \item[] Answer: \answerNo{} 
    \item[] Justification: 
    Alghough we didn't provide the statistical metrics, we have run extensive experiments on large-scale datasets like ImageNet to demonstrate the statistical significance.

\item {\bf Experiments compute resources}
    \item[] Question: For each experiment, does the paper provide sufficient information on the computer resources (type of compute workers, memory, time of execution) needed to reproduce the experiments?
    \item[] Answer: \answerYes{} 
    \item[] Justification: 
    We have provide sufficient information on computer resources to reproduce the results.
    
\item {\bf Code of ethics}
    \item[] Question: Does the research conducted in the paper conform, in every respect, with the NeurIPS Code of Ethics \url{https://neurips.cc/public/EthicsGuidelines}?
    \item[] Answer: \answerYes{} 
    \item[] Justification: 
    The research conducted in this paper totally confirm the NeurlPS Code of Ethics.

\item {\bf Broader impacts}
    \item[] Question: Does the paper discuss both potential positive societal impacts and negative societal impacts of the work performed?
    \item[] Answer: \answerYes{} 
    \item[] Justification:
    This paper provide an extra impact statement section in the appendix to discuss the societal impacts.
    
\item {\bf Safeguards}
    \item[] Question: Does the paper describe safeguards that have been put in place for responsible release of data or models that have a high risk for misuse (e.g., pre-trained language models, image generators, or scraped datasets)?
    \item[] Answer: \answerNA{} 
    \item[] Justification: 
    This paper doesn't include data or models that have a high risk of misuse.

\item {\bf Licenses for existing assets}
    \item[] Question: Are the creators or original owners of assets (e.g., code, data, models), used in the paper, properly credited and are the license and terms of use explicitly mentioned and properly respected?
    \item[] Answer: \answerYes{} 
    \item[] Justification: 
    All the data and models in the paper are publicly available.

\item {\bf New assets}
    \item[] Question: Are new assets introduced in the paper well documented and is the documentation provided alongside the assets?
    \item[] Answer: \answerNA{} 
    \item[] Justification: 
    This paper doesn't include any new assets.

\item {\bf Crowdsourcing and research with human subjects}
    \item[] Question: For crowdsourcing experiments and research with human subjects, does the paper include the full text of instructions given to participants and screenshots, if applicable, as well as details about compensation (if any)? 
    \item[] Answer: 
    \answerNA{}
    \item[] Justification:
    This paper doesn't involve any crowdsourcing experiments and research with human subjects.

\item {\bf Institutional review board (IRB) approvals or equivalent for research with human subjects}
    \item[] Question: Does the paper describe potential risks incurred by study participants, whether such risks were disclosed to the subjects, and whether Institutional Review Board (IRB) approvals (or an equivalent approval/review based on the requirements of your country or institution) were obtained?
    \item[] Answer: \answerNA{} 
    \item[] Justification: 
    This paper doesn't include any crowdsourcing nor research with human subjects.

\item {\bf Declaration of LLM usage}
    \item[] Question: Does the paper describe the usage of LLMs if it is an important, original, or non-standard component of the core methods in this research? Note that if the LLM is used only for writing, editing, or formatting purposes and does \emph{not} impact the core methodology, scientific rigor, or originality of the research, declaration is not required.
    \item[] Answer: \answerNA{} 
    \item[] Justification: The core method in this paper does not involve LLMs as any important, original, or non-standard components.

\end{enumerate}

\end{document}